\documentclass{article}

\usepackage[accepted]{icml2025}
\makeatletter
\renewcommand{\Notice@String}{Copyright 2026, Accenture. All   
  rights reserved.}  
\makeatother

\usepackage{graphicx}
\usepackage{booktabs}
\usepackage{amsmath}
\usepackage{amssymb}
\usepackage{mathtools}
\usepackage{textcomp}
\usepackage{multirow}
\usepackage{algorithm}
\usepackage{algorithmic}
\usepackage{enumitem}
\usepackage{listings}
\usepackage[breakable]{tcolorbox}

\usepackage{tikz}
\usetikzlibrary{shapes.geometric, arrows.meta, positioning, fit, backgrounds, calc}
\usepackage{xcolor}

\usepackage{orcidlink}

\icmltitlerunning{VANGUARD: Video Anomaly Understanding through Reasoning and Grounding}

\begin{document}

\twocolumn[
\icmltitle{Reasoning-Guided Grounding: Elevating Video Anomaly Detection through Multimodal Large Language Models}

\begin{icmlauthorlist}
\icmlauthor{Sakshi Agarwal}{accenture}
\icmlauthor{Aishik Konwer}{accenture}
\icmlauthor{Ankit Parag Shah}{accenture}
\end{icmlauthorlist}

\icmlaffiliation{accenture}{Center of Advanced AI, Accenture}

\icmlcorrespondingauthor{Sakshi Agarwal}{sakshi.aa.agarwal@accenture.com}
\icmlcorrespondingauthor{Aishik Konwer}{aishik.konwer@accenture.com}
\icmlcorrespondingauthor{Ankit Parag Shah}{ankit.parag.shah@accenture.com}

\icmlkeywords{Video Anomaly Detection, Vision-Language Models, Spatial Grounding, Explainable AI, Chain-of-Thought Reasoning}

\vskip 0.1in
]

\printAffiliationsAndNotice{}

% ============================================
% ABSTRACT
% ============================================
\begin{abstract}
% Abstract Section
% This file contains the paper abstract

Video Anomaly Detection (VAD) has traditionally been framed as binary classification or outlier detection, providing neither interpretable reasoning nor precise spatial localization of anomalous events. While Vision-Language Models (VLMs) offer rich scene understanding, they struggle with reliable spatial grounding---often producing hallucinated or geometrically invalid bounding boxes when asked to localize objects. We propose \textsc{Vanguard} (\textbf{V}ideo \textbf{A}nomaly u\textbf{N}derstandin\textbf{G} thro\textbf{U}gh re\textbf{A}soning and g\textbf{R}oun\textbf{D}ing), a framework that unifies anomaly classification,  spatial grounding and chain-of-thought reasoning within a single VLM. \textsc{Vanguard} introduces a three-stage curriculum that progressively layers training objectives: (1)~classifier warmup on frozen backbone features, (2)~LoRA-adapted spatial grounding, and (3)~chain-of-thought generation.
To overcome the sparse annotation typical of VAD benchmarks, we employ a teacher-student annotation pipeline in which a VLM (Qwen3-VL-4B) generates structured per-subclip reasoning trajectories based on manual annotations available from UCA Dataset. Further, GroundingDINO provides bounding box supervision. On UCF-Crime, \textsc{Vanguard} achieves $94\%$ ROC-AUC with $84\%$ F1 while  simultaneously producing interpretable CoT explanations and spatial grounding of anomalous objects---capabilities absent from prior VAD methods. Ablations confirm that staged training outperforms monolithic optimization, and that structured reasoning acts as an implicit regularizer yielding more balanced predictions than classification-only fine-tuning. Zero-shot transfer to XD-Violence and ShanghaiTech demonstrates cross-domain generalization without target-domain adaptation.
\end{abstract}

% ============================================
% MAIN CONTENT
% ============================================

% Introduction Section
\section{Introduction}
\label{sec:introduction}

Video Anomaly Detection (VAD) represents a critical challenge in computer vision with far-reaching applications in public safety, industrial monitoring, and autonomous systems. The fundamental goal of identifying events that deviate from expected patterns has driven decades of research, yet existing approaches suffer from two persistent limitations: \textit{lack of interpretability} and \textit{absence of precise spatial localization}.

\paragraph{The Current Landscape}

Traditional VAD methods treat the problem as binary classification or one-class novelty detection \cite{sultani2018real}. These approaches, while achieving reasonable detection accuracy measured by Area Under the ROC Curve (AUC), provide only video-level or frame-level labels. When a surveillance system flags an anomaly, operators receive no explanation of \textit{what} is anomalous or \textit{where} to look. This opacity creates a fundamental barrier to deployment in safety-critical applications where human oversight remains essential.

Recent advances in Vision-Language Models (VLMs) have opened new possibilities. Models such as Qwen-VL \cite{bai2023qwen, qwenvl3}, LLaVA \cite{liu2024visual}, and GPT-4V \cite{gpt4v} have demonstrated remarkable capabilities in scene understanding, visual question answering, and natural language generation. Several works have begun exploring VLMs for anomaly detection. ASK-HINT \cite{askhint} designs fine-grained, action-centric prompts to unlock reasoning in frozen VLMs, while VERA \cite{ye2025vera} automates this prompt discovery process and optimizes these natural-language guiding questions through data-driven verbal interactions between learner and optimizer VLMs. On the other hand, instead of prompt-finetuning, LAVAD \cite{zanella2024harnessing} generates per-frame captions with a VLM and delegates anomaly scoring to an LLM. Apart from computational time requirements to caption every frame, LAVAD suffers from a major drawback that is based solely on the textual caption stream without direct access to visual cues. These methods keep the VLM frozen and rely entirely on prompt design, limiting their ability to learn discriminative visual representations from training data.  

\paragraph{The Grounding Gap} 
Despite these advances, a critical gap remains: \textbf{VLMs consistently fail at precise spatial grounding in anomaly detection contexts}. Our preliminary experiments with state-of-the-art VLMs reveal that while classification accuracy and reasoning quality can be satisfactory, bounding box predictions are: (1)~\textbf{Hallucinated}: boxes reference non-existent objects; (2)~\textbf{Imprecise}: coordinates deviate significantly from actual anomaly locations; and (3)~\textbf{Inconsistent}: predictions vary wildly across similar frames. This ``Bounding Box Failure'' is not merely a minor limitation---it fundamentally undermines the utility of VLM-based approaches for practical surveillance applications. An explainable anomaly detector that cannot reliably point to \textit{where} the anomaly occurs provides incomplete actionable intelligence.

We argue that effective VLM-based anomaly detection requires all three capabilities simultaneously: (i)~reliable binary classification, (ii)~interpretable chain-of-thought reasoning~\cite{wei2022chain} that explains the decision, and (iii)~spatial grounding that localizes anomalous objects with bounding boxes. Achieving all three within a single model presents distinct challenges. Na\"ively fine-tuning a VLM with all objectives at once leads to competing gradients and training instability, as randomly initialized heads produce noisy supervision signals that corrupt the pretrained backbone.  

% We propose \textsc{Vanguard} (\textbf{V}ideo \textbf{A}nomaly u\textbf{N}derstandin\textbf{G} thro\textbf{U}gh re\textbf{A}soning and g\textbf{R}oun\textbf{D}ing), a framework that unifies all three capabilities within a single VLM through a structured curriculum. A central challenge is that existing VAD benchmarks provide only video-level binary labels for long, untrimmed surveillance footage---far too coarse for training a model that must reason about specific objects and their locations. We address this through an automated annotation pipeline: each video is segmented into semantically coherent subclips via CLIP-based keyframe detection. For these subclips, we use temporal subclip descriptions from UCA Dataset \cite{ucadataset} to generate structured per-subclip annotations with object-level event labels and free-text reasoning using Qwen3-VL-4B. GroundingDINO then provides bounding box supervision for each annotated object. This transforms ${\sim}$800 weakly-labeled training videos into ${\sim}$30{,}000 richly-annotated subclip samples, each with object identities, anomaly reasoning, spatial grounding, and DPO preference pairs---without any manual annotation.

\subsection{Our Approach: VANGUARD}

We propose \textsc{Vanguard} (\textbf{V}ideo \textbf{A}nomaly u\textbf{N}derstandin\textbf{G} thro\textbf{U}gh re\textbf{A}soning and g\textbf{R}oun\textbf{D}ing), a framework that unifies all three capabilities within a single VLM through a structured curriculum. A central challenge is that existing VAD benchmarks provide only video-level binary labels for long, untrimmed surveillance footage---far too coarse for training a model that must reason about specific objects and their locations. We address this through an automated annotation pipeline: each video is segmented into semantically coherent subclips via CLIP-based keyframe detection, Qwen3-VL-4B generates structured per-subclip annotations with object-level event labels and free-text reasoning, and GroundingDINO provides bounding box supervision for each annotated object. This transforms ${\sim}$1000 weakly-labeled training videos into ${\sim}$40{,}000 richly-annotated subclip samples, each with object identities, anomaly reasoning, and spatial grounding ---without any manual annotation.

\textsc{Vanguard} builds on a frozen Qwen3-VL-4B backbone adapted with LoRA and augmented with task-specific head for classification. Training proceeds in three stages: (1)~\emph{classifier warmup}, where only the classifier head is trained on stable backbone features; (2)~\emph{spatial grounding}, where LoRA adapters are unfrozen and a dedicated spatial loss supervises the localization of anomalous objects; and (3)~\emph{chain-of-thought fine-tuning}, where the model learns to generate structured reasoning.

Our contributions are as follows:
\begin{enumerate}
    \item We introduce \textsc{Vanguard}, the first VAD framework that jointly produces anomaly classification, chain-of-thought explanations, and spatial grounding of anomalous objects within a single VLM.
    \item We present an automated annotation pipeline that transforms weakly-labeled surveillance videos into richly-annotated subclip samples---with object-level event tags, free-text reasoning, and spatial bounding boxes ---using the same VLM, temporal annotations from UCA Dataset, and an open-vocabulary detector, requiring no manual annotation.
    \item We propose a three-stage curriculum that progressively layers three training objectives, resolving the gradient conflicts that arise from simultaneous multi-task optimization. We show that staged training outperforms monolithic optimization and that structured reasoning acts as an implicit regularizer.
    \item We provide comprehensive evaluation demonstrating state-of-the-art performance on UCF-Crime, XD-Violence and ShanghaiTech, with the \textbf{first reported spatial grounding metrics (bounding-box IoU)} for video anomaly detection---a new benchmark for future research.
    % \item We evaluate on UCF-Crime, XD-Violence, and ShanghaiTech, demonstrating \todo{competitive/state-of-the-art} classification accuracy alongside interpretable explanations and spatial grounding---capabilities absent from prior VAD methods.
\end{enumerate}

The remainder of this paper is organized as follows. Section~\ref{sec:related_work} reviews related work in video anomaly detection, VLMs, and spatial grounding. Section~\ref{sec:method} details our proposed methodology. Section~\ref{sec:experiments} presents experimental results and analysis. Section~\ref{sec:conclusion} concludes with discussion and future directions.

% Related Work Section
\section{Related Work}
\label{sec:related_work}

\begin{figure*}[t] \centering  \includegraphics[width=\textwidth]{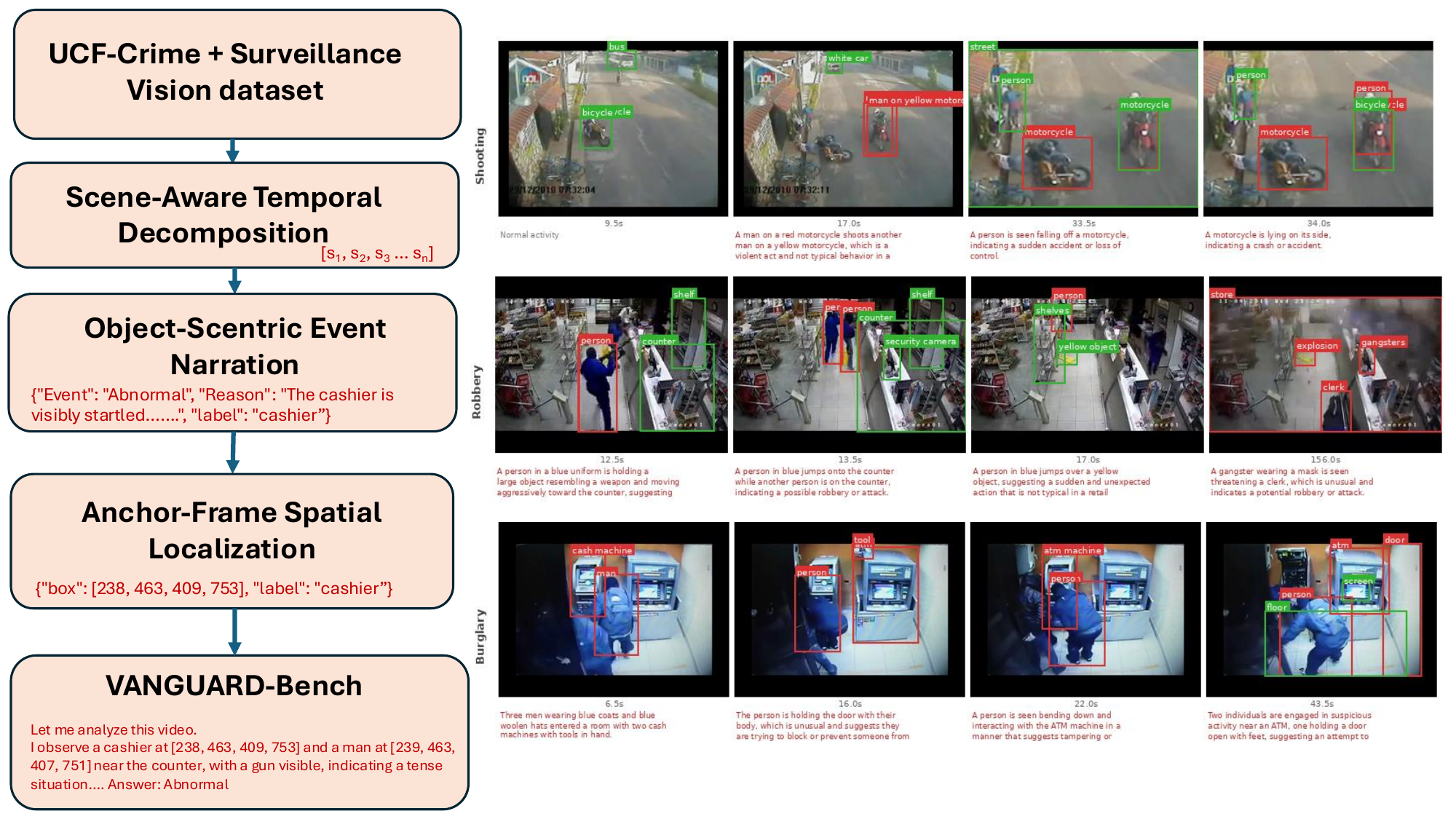} \caption{\textbf{VANGUARD-Bench dataset creation pipeline and sample annotations.} \textit{Left:} Starting from UCF-Crime surveillance videos, we extract CLIP-based keyframes to segment each video into temporally distinct subclips. A vision-language model (Qwen3-VL)  generates per-object annotations with event classification (\textit{Normal}/\textit{Abnormal}) and natural language reasoning. GroundingDINO then localizes each object with bounding boxes via Hungarian matching for one-to-one assignment. \textit{Right:} Example annotations across three anomaly types. Each column shows the  grounding frame of a subclip with spatially grounded objects ({\color{red}red} = abnormal, {\color{green}green} = normal) and the VLM-generated reasoning below. The pipeline produces  structured \{Event, Reason, label, box\} annotations that enable joint classification and spatial grounding supervision.}
 \label{fig:vanguard-dataset}
\end{figure*}

\paragraph{Traditional Methods}
VAD has evolved from unsupervised one-class approaches~\cite{hasan2016learning, park2020learning, gong2019memorizing} that learn normality patterns from unlabeled data, to weakly-supervised methods that leverage video-level labels for learning~\cite{sultani2018real, tian2021weakly, chen2023mgfn, zhou2023urdmu}. While these approaches have improved accuracy on benchmarks like UCF-Crime through robust temporal feature learning~\cite{tian2021weakly}, contrastive objectives~\cite{chen2023mgfn}, and memory-based designs~\cite{zhou2023urdmu}, they remain limited to binary anomaly scoring without fine-grained classification or explanation of the detected behaviors~\cite{tian2021weakly, Li_Liu_Jiao_2022}. 

\paragraph{CLIP-based approaches.} Vision-language representations have been adapted for fine-grained anomaly classification by aligning video features with textual anomaly descriptions~\cite{wu2023vadclip, pu2024learningpromptenhancedcontextfeatures, Kim_Yoon_Choi_Sull_2023}. Building on learnable context vectors for prompt tuning~\cite{Zhou_2022}, early approaches employ trainable textual templates to describe anomaly categories~\cite{wu2023vadclip}, but remain sensitive to template design. Subsequent work mitigates this by automatically deriving prompts from external knowledge bases, reducing manual effort while improving fine-grained anomaly discrimination~\cite{pu2024learningpromptenhancedcontextfeatures}.  

%\paragraph{Learning fine-grained anomalies} Ex‑VAD \cite{pmlr-v267-huang25ad} leverages a VLM to extract frame‑level captions and an LLM to generate enriched video‑level explanations, labels, and anomaly‑specific sentences. It further integrates textual and visual representations derived from CLIP to train a unified classifier using binary and multi‑class objectives alongside a contrastive loss for robust fine‑grained anomaly detection. 
% \cite{du2024ecva} introduces a chain‑of‑thought–based prompting strategy and trains a connector module within a vision–language model to align visual and textual representations for catching fine-grained anomalies. %It also proposes AnomEval, a metric for evaluating causal reasoning in video anomaly models, and provides the ECVA benchmark dataset, which we use to evaluate our method.
% to generate explanations as additional context to a VLM detector. 

\paragraph{MLLMs and Reasoning for VAD.}       Vision-language models have opened new paradigms for VAD by enabling natural language reasoning over video content. One line of work structures prompts into semantically coherent groups and formulates fine-grained guiding questions that align model predictions with visual cues (ASK-Hint~\cite{askhint}), with subsequent methods optimizing such guiding questions through verbalized learning without fine-tuning or instruction-data tuning (VERA,~\cite{ye2025vera}). A complementary direction pursues entirely training-free detection: inducing normality rules from few-shot normal references and flagging test frames that violate them~\cite{yang2024anomalyruler}, or generating frame-level captions with a VLM and delegating anomaly reasoning to an LLM~\cite{zanella2024harnessing, pmlr-v267-huang25ad}. Beyond detection, recent work extends to temporal grounding, localizing \emph{when} anomalous events occur through dynamic textual prompt guidance~\cite{vagu2025}.                           

%VLAVAD~\cite{li2024vlavad} introduces a Selective Prompt Adapter for semantic anomaly analysis, 

A separate line of work fine-tunes VLMs to produce natural language explanations alongside anomaly predictions. Multimodal instruction-tuning datasets have been curated at scale to train MLLMs with learned temporal samplers for explainable detection~\cite{zhang2024holmesvad}, while explicit motion-branch guidance enables interactive open-world anomaly understanding through both description generation and question answering~\cite{tang2024hawk}. More recently, reinforcement learning with self-verification has been applied to produce structured, perception-to-cognition chain-of-thought anomaly explanations~\cite{huang2025vadr1}. While these methods advance interpretability, none achieve reliable spatial grounding---the core contribution of our work.   

%% Sakshi : Introduce these two datasets only if you plan to benchmark on them : VAGU~\cite{vagu2025} introduces the first benchmark jointly addressing anomaly grounding and understanding. 

\paragraph{Spatial Grounding in VLMs.} Grounding language to image regions is a fundamental multimodal challenge, addressed through diverse strategies: pixel-wise segmentation mask prediction via dedicated decoders~\cite{rasheed2024glamm}, location tokens interleaved in the text sequence~\cite{peng2023kosmos}, or bounding box coordinates and metric distances represented directly as text~\cite{chen2023shikra, bai2025qwen25vltechnicalreport, chen2024spatialvlm, qwenvl3}. Recent work further shows that such grounding abilities can be added post-hoc to pre-trained VLMs through forget-free tuning, and that synthetic spatially-grounded chain-of-thought data can teach models to reason step-by-step through explicit bounding box intermediate steps~\cite{Bhowmik_2025_ICCV, visualcot}. 

We adopt Qwen3-VL-4B~\cite{qwenvl3} as our vision--language backbone, which natively supports coordinate bin tokens for spatial referencing. However, we find that these tokens alone are
   insufficient for VAD---without targeted multi-objective training, the model fails to produce reliable spatial predictions for anomalous objects. This observation motivates VANGUARD,   
  which brings spatial grounding into the video anomaly detection setting to localise \textit{anomalous} objects rather than arbitrary referents. Overall, existing grounding methods largely overlook the fine-grained human-object interactions and action semantics that define complex anomalies in surveillance video. \textsc{Vanguard} is, to our knowledge, the first VAD framework to jointly deliver reliable classification, chain-of-thought explanations, and bounding-box localization within a single fine-tuned VLM.

\section{VANGUARD-Bench}
\label{sec:benchmark}
In this section, we detail the construction of the VANGUARD-Bench dataset through a hierarchical annotation pipeline that progressively enriches raw surveillance videos at three levels of increasing granularity. At the coarsest level, we decompose each video temporally into scene-coherent subclips. At the intermediate level, we narrate each subclip by identifying individual objects and their roles in normal or anomalous events. At the finest level, we localize each narrated object spatially within the frame. This coarse-to-fine hierarchy---\textit{video}~$\rightarrow$~\textit{subclip}~$\rightarrow$~\textit{object}~$\rightarrow$~\textit{bounding box}---mirrors the easy-to-hard principle of curriculum learning~\cite{bengio2009curriculum} and produces annotations that jointly capture the temporal, semantic, and spatial dimensions of anomalous events, enabling the model to learn \textit{when}, \textit{what}, \textit{who}, and \textit{where} an anomaly occurs. The overall pipeline of the data engine is shown in Fig.~\ref{fig:vanguard-dataset}.

\subsection{Data Collection}

Following Holmes-VAD~\cite{zhang2024holmesvad}, we source our raw videos from UCF-Crime~\cite{ucf_crime}, %and XD-Violence~\cite{xdviolence},
the largest weakly-supervised VAD benchmark, whose scale substantially surpasses other available datasets~\cite{Wang2010AnomalyDI, Lu_2013_ICCV, Luo_2017_ICCV} and whose video-level labels provide a reliable starting point for our annotation pipeline. We also utilize the existing SurveillanceVision dataset \cite{ucadataset} that contains manual annotations of the UCF-Crime dataset. Following a manual quality review to remove corrupted and heavily occluded clips, we retain $1,610$ untrimmed videos: $810$ abnormal and $800$ normal from the training-split in UCF-Crime. %, and 1\,905 abnormal and 2\,032 normal from XD-Violence.

% We first collect videos from the training sets of the two largest weakly-supervised VAD datasets,
% UCF-Crime \cite{ucf_crime} and XD-Violence \cite{xdviolence}, because their video quantity far exceeds that of other existing
% datasets \cite{Wang2010AnomalyDI, Lu_2013_ICCV, Luo_2017_ICCV}, and their video-level annotations provide a solid foundation for further data processing. After filtering out some low-quality videos via human inspection, we collected a total of 5547 untrimmed videos, include 810/800 abnormal/normal videos from UCF-Crime and 1905/2032 abnormal/normal videos from XD-Violence. 

\subsection{Hierarchical Annotation Enhancement}
\paragraph{Scene-Aware Temporal Decomposition}
The first level of our hierarchy partitions each video into visually distinct subclips along the temporal axis. Rather than splitting at fixed intervals as done in the
  existing SurveillanceVision dataset~\cite{ucadataset}, we perform \textit{scene-aware} subclip segmentation using CLIP ViT-B/32~\cite{radford2021clip}: we compute image
  embeddings for every 15th frame and declare a subclip boundary whenever the cosine similarity to the previous boundary frame drops below a threshold ($0.92$), indicating a     
  visual scene change. Each subclip captures a temporally coherent segment of the video, indexed by its start and end frame numbers to preserve temporal position within the
  original video. Across $1,610$ training videos, this process yields a total of $39,003$ candidate subclips (mean 22.5, median 7 per video). From this set, we retain up to two per video, resulting in a total of $2,220$ subclips, one from each class %by prioritizing subclips whose annotated objects include anomalous events 
  and backfilling remaining slots with uniformly sampled normal subclips.

 %To keep training tractable, we retain at most two subclips per video, prioritizing those containing anomalous events. Across 1\,288 training videos, this yields 2\,092 subclips.
\paragraph{Object-centric Event Narration}
The second level moves from subclips down to individual objects. For this step, we use the existing SurveillanceVision dataset~\cite{ucadataset} and align annotations to each subclip by timestamp matching. Rather than assigning a single label or detailed captions per subclip as performed in~\cite{zhang2024holmesvad, huang2025vadr1}, we follow the model-generated annotation paradigm~\cite{wang2023selfinstruct} and prompt a vision--language model (Qwen3-VL~\cite{qwenvl3}) with each subclip's frames and aligned temporal sentences to enumerate every salient object. For each object, the VLM produces a structured \textit{event narration}: (i)~an \textit{Event} tag (Normal/Abnormal), (ii)~a natural-language \textit{Reason} grounded in visual evidence, (iii)~the object \textit{label}, and (iv)~a \textit{confidence} score. This object-centric scheme disentangles the contributions of different actors and artifacts, capturing \textit{what} anomaly occurs, \textit{which} objects are responsible, and \textit{why}. This results in a total of $159,008$ objects labeled across the videos (mean 4.1 objects per subclip).

\paragraph{Anchor-Frame Spatial Localization}
The third level grounds each narrated object in pixel space. Following the bipartite matching strategy of DETR~\cite{carion2020end}, we query GroundingDINO~\cite{liu2024grounding} with each object's label on the \textit{last frame} of the subclip, which serves as the anchor frame capturing the culminating state where anomalous actions are most fully manifest. The highest-confidence detection is retained as the bounding box $[x_1, y_1, x_2, y_2]$. %Scene-level detections covering more than 50\% of the frame area are filtered out, as labels such as building'' or road'' do not provide useful spatial grounding. 
This yields spatially grounded annotations for $147,067$ of $159,008$ objects ($92.5\%$) across all subclips.

\subsection{Constructing VANGUARD-Bench}
The hierarchical annotation pipeline described above produces rich per-subclip metadata—object identities, event labels, causal reasons, and bounding boxes—but does not yet cast this information into the conversational format required for instruction tuning~\cite{wei2022finetuned}. To bridge this gap, we aggregate subclip-level annotations across the full video timeline and utilize
  Qwen3-VL~\cite{qwenvl3}—chosen for its native video understanding and open-source availability—to synthesize subclip-level, spatially grounded instruction data. 
  For each subclip $s$, we aggregate the object annotations—event tags ${y_k}$, causal reasons ${r_k}$, and bounding box coordinates ${b_k}$ localized on each subclip's last frame—into a unified spatiotemporal context $\mathcal{C}_v$ that spans the full subclip timeline. We then design a task prompt $P_t$ that instructs the VLM to produce a chain-of-thought    
  analysis~\cite{wei2022chain} referencing explicit bounding box coordinates for every mentioned object in the last frame of the subclip. The \textit{full subclip  } is passed as visual input alongside $\mathcal{C}_s$
   and $P_t$, enabling the model to leverage motion cues and inter-frame dynamics invisible in static snapshots. The generated response is paired with an anomaly-aware question $P_d$ to form a
 subclip-level instruction item:
  \begin{equation}
    \mathcal{I}_s = {\texttt{user}: P_d,;\texttt{assistant}: \text{VLM}(P_t, s, \mathcal{C}_s)}
  \end{equation}
  Each response is post-processed to prepend a deterministic detection block that preserves spatial coordinates verbatim, followed by the VLM's temporally coherent analysis, ensuring that bounding boxes are faithfully retained regardless of paraphrasing.
\section{Method}
\label{sec:method}
\begin{figure*}[t]
\centering
\includegraphics[width=\textwidth]{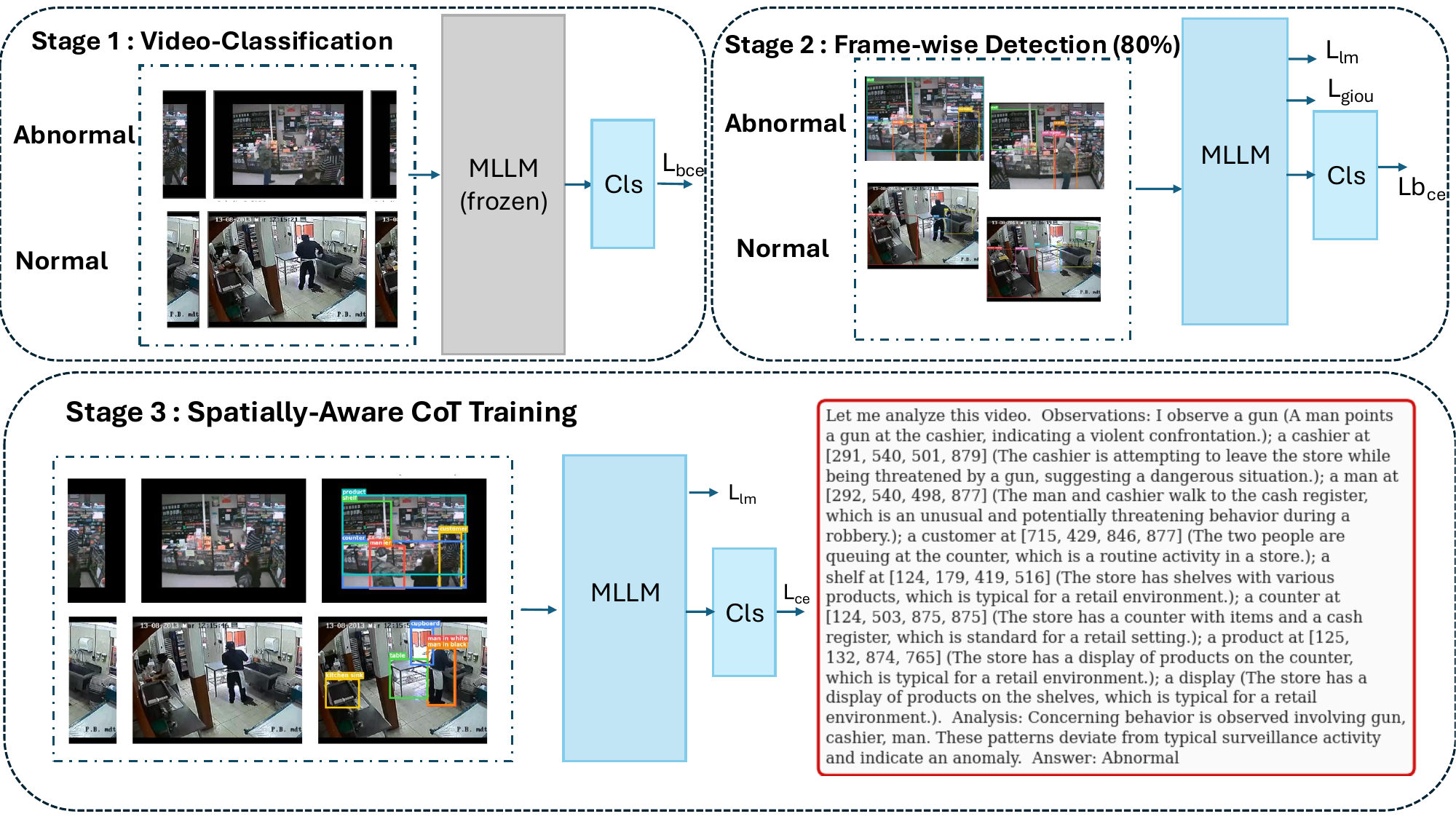}\caption{
Overview of \textsc{Vanguard}'s three-stage curriculum training procedure. Stage~1 trains only the classification head on video-level labels using $\mathcal{L}_\text{bce}$. Stage~2 unfreezes LoRA adapters and introduces mixed data (80\% image-level detection, 20\% video-level CoT) with spatial grounding losses $\mathcal{L}_\text{lm}$ and $\mathcal{L}_\text{giou}$. Stage~3 refines reasoning on video-level CoT data only, dropping the GIoU loss. Each stage loads the checkpoint from the previous one, progressively building from classification to spatial grounding to chain-of-thought reasoning.
}
\label{fig:curriculum}
\end{figure*}

Leveraging the proposed VANGUARD-Bench dataset, we present VANGUARD, an interpretable video anomaly detection framework that jointly produces (i)~a binary anomaly classification, (ii)~a natural-language chain-of-thought (CoT) explanation~\cite{wei2022chain} grounded in temporal context, and (iii)~spatial localization of anomalous objects via bounding boxes. Figure~\ref{fig:curriculum} illustrates the overall training procedure. Below we detail the model architecture (\S\ref{sec:architecture}), multi-objective training losses and the curriculum learning strategy (\S\ref{sec:curriculum}).

%% ----------------------------------------------------------------

%% ----------------------------------------------------------------
\subsection{Model Architecture}
\label{sec:architecture}

We pick Qwen3-VL-4B-Instruct~\cite{qwenvl3} as our vision-language backbone, adapted via LoRA~\cite{hu2022lora}.  %The model processes interleaved video and text tokens; video frames are patchified and merged with a spatial merge size of $2$, producing a token grid of shape $T \times \lfloor H/2 \rfloor \times \lfloor W/2 \rfloor$ per video. 
We wrap the base model with a classifier head for a binary prediction of normal/abnormal video. The classifier consists of a two-layer MLP %($2560 \to 2560 \to 128$, ReLU activation) 
that maps the pooled hidden state from the last layer of the base model to a 128-dimensional feature space, followed by a Dropout ($p{=}0.5$) and a linear layer ($128 \to 1$) that produces a binary anomaly logit $l$. %from the projected features. % 3.  \textbf{Object head}: Dropout %($p{=}0.3$) followed by a linear layer ($128 \to 16$) predicting 16 object categories (see \S\ref{sec:obj_aux}); 4. \textbf{Object-conditioned bin head}: we introduce per-category learnable embeddings, and add them to the shared feature projection, h. This ensures that each category gets the same scene features and its own identity bias. A single MLP then decodes each combined vector into 4 coordinate distributions, where each coordinate $(x1, y1, x2, y2)$ is a $1001$-way classification over bins. %  Predicts bounding box coordinates for each object category via a  per-category learnable embeddings, and a box decoder (see \S\ref{sec:binned}).

\textbf{Note:} Qwen3-VL natively includes coordinate tokens for general grounding. Our contribution is demonstrating that these tokens fail for VAD without proper training, and developing the multi-objective loss weighting that enables reliable spatial prediction. For models without native coordinate tokens (e.g., LLaVA), a vocabulary extension with spatial bin tokens would be necessary.
% \paragraph{Pooling with spatial attention guidance.}
% \label{sec:spatial_pool}
% The pooled representation is computed as a weighted mean over the sequence:
% \begin{equation}
%     \mathbf{h} = \frac{\sum_{i} w_i \cdot m_i \cdot \mathbf{h}_i}{\sum_{i} w_i \cdot m_i},
%     \label{eq:pool}
% \end{equation}
% where $\mathbf{h}_i$ is the hidden state at position $i$, $m_i$ is the attention mask, and $w_i$ is a spatial weight. For text tokens, $w_i{=}1$. For video tokens whose spatial centre falls inside an annotated abnormal-object bounding box, $w_i{=}\alpha$ (default $\alpha{=}3.0$); all other video tokens receive $w_i{=}1$. Token centres are computed from the merged video grid: $h_\text{norm} = (h_\text{idx} + 0.5) / \lfloor H/2\rfloor$, $w_\text{norm} = (w_\text{idx} + 0.5) / \lfloor W/2 \rfloor$, and a token is considered inside a box $[x_1, y_1, x_2, y_2]$ if $x_1 \le w_\text{norm} \le x_2$ and $y_1 \le h_\text{norm} \le y_2$.

% At test time, no bounding boxes are available and all weights reduce to $w_i{=}1$ (uniform pooling). To bridge this train--test gap, we anneal $\alpha$ linearly from its initial value to $1.0$ over training steps (see \S\ref{sec:anneal}) and add an attention alignment loss (see \S\ref{sec:align}).

%% ----------------------------------------------------------------

\begin{table*}[t]
      \centering
      \small
      \setlength{\tabcolsep}{4pt}
      \resizebox{\textwidth}{!}{%
      \begin{tabular}{lcccc|cc|cc}
      \toprule
      \multirow{2}{*}{Method} &
      \multicolumn{4}{c}{UCF-Crime} &
      \multicolumn{2}{c}{XD-Violence} &
      \multicolumn{2}{c}{ShanghaiTech Campus} \\
      \cmidrule(lr){2-5} \cmidrule(lr){6-7} \cmidrule(lr){8-9}
      & AUC $\uparrow$ & F1 $\uparrow$ & meanIoU $\uparrow$ & R@25 $\uparrow$
      & AUC $\uparrow$ & F1 $\uparrow$
      & AUC $\uparrow$ & F1 $\uparrow$ \\
      \midrule

      ASK-Hint & 0.8983 & -- & -- & -- & 0.9031 & -- & -- & -- \\

      VERA & 0.8655 & -- & -- & -- & 0.8826 & -- & -- & -- \\

      LaVAD & 0.8028 & -- & -- & -- & 0.8536 & -- & -- & -- \\

      Flashback (2025) & 0.8729 & -- & -- & -- & 0.9054 & -- & -- & -- \\
      \midrule

      HolmesVAD & 0.7299 & 0.3721 & -- & -- & \textbf{0.9588} & 0.8000 & \underline{0.5000}$^\star$ &
      0.0000$^\star$ \\

      VADR1 & 0.8445 & 0.8506 & -- & -- & 0.9158 & \textbf{0.9455} & \textbf{0.5158}$^\star$ &
      \textbf{0.1667}$^\star$ \\
      \midrule

      Kosmos-2 & 0.6709 & 0.6879 & 0.00 & 0.00 & 0.3505 & 0.7419 & 0.4329$^\star$ & -- \\

      Visual-CoT & 0.5827 & 0.6650 & \underline{0.45} & \textbf{0.74} & 0.3453 & 0.7682 & 0.4404$^\star$ & -- \\

      SpatialVLM & 0.8318 & 0.4783 & 0.32 & 0.44 & \underline{0.9475} & 0.7748 & 0.4945$^\star$ & -- \\
      \midrule

      \textsc{Vanguard} (Stage-1) & \textbf{0.9436} & \underline{0.7750} & -- & -- & 0.9135 & \underline{0.8571} & 0.4085$^\star$ & 0.0000$^\star$ \\
      \textsc{Vanguard} & \underline{0.9378} & \textbf{0.8360} &\textbf{0.62} & \underline{0.51} & 0.9149 & 0.8403 & 0.4732$^\star$ &
  \underline{0.1215}$^\star$ \\
      \bottomrule
      \end{tabular}%
      }
      \caption{Comparison with recent video anomaly detection methods on
      UCF-Crime, XD-Violence, and ShanghaiTech Campus datasets.
      We report detection metrics (AUC, F1) and spatial localization metrics
      (meanIoU, R@25).
      $\uparrow$ indicates higher is better. $\star$ denotes frame-level
      evaluation.}
      \label{tab:vad_results}
\end{table*}

  \subsection{Learning Strategy}
  \label{sec:curriculum}

  Training all objectives simultaneously from the start forces the model to balance competing gradients before any single objective has converged~\cite{sener2018multitask, kendall2018multitask}. We first define
   the individual loss functions, then describe the curriculum that introduces them gradually across three stages.                              
  \subsubsection{Loss Functions.}

  \textit{Binary cross-entropy ($\mathcal{L}_\text{bce}$).} The classification head produces a logit $\ell_i$ for each input, supervised by:
  \begin{equation}
      \mathcal{L}_\text{bce} = -\frac{1}{N}\sum_{i}\left[y_i \log \sigma(\ell_i) + (1{-}y_i)\log(1{-}\sigma(\ell_i))\right].
  \end{equation}

  \textit{Text generation ($\mathcal{L}_\text{lm}$).} The standard autoregressive cross-entropy from the base model's LM head, computed over assistant response tokens only (prompt tokens are masked):
  \begin{equation}
      \mathcal{L}_\text{lm} = -\sum_{t \in \mathcal{A}} \log p_\theta(x_t \mid x_{<t}),
  \end{equation}
  where $\mathcal{A}$ is the set of assistant response token positions, $x_t$ is the ground-truth token at position $t$, and $x_{<t}$ denotes all preceding tokens. The target text varies by training stage: JSON
   detection output for image-level samples and temporal chain-of-thought with inline bounding boxes for video-level samples. However, $\mathcal{L}_\text{lm}$ treats each digit token as an independent
  classification over the vocabulary—a coordinate prediction of 500 when the target is 501 is penalized identically to a prediction of 100—making it insufficient for accurate bounding box regression on its own.

  \textit{Text-coordinate GIoU ($\mathcal{L}_\text{giou}$).}
  \label{sec:text_coord_giou}
  To address this, we introduce a geometry-aware regression loss on bounding box coordinates directly through the language model's token predictions, without any separate box head. At each coordinate digit
  position in the teacher-forced sequence, the model's next-token logits are restricted to digit tokens (0--9) and softmaxed to produce a differentiable ``soft'' coordinate value, following integral
  regression~\cite{sun2018integral, nibali2018numerical, chen2022pix2seq}. Four such values form a predicted box $\hat{\mathbf{b}} \in [0,1]^4$. Ground-truth boxes are parsed from the target response, and
  predicted and ground-truth boxes are matched by object label:
  \begin{equation}
      \mathcal{L}\text{giou} = \frac{1}{M}\sum_{m=1}^{M}\left(1 - \text{GIoU}(\hat{\mathbf{b}}_m, \mathbf{b}_m)\right),
  \end{equation}
  where $M$ is the number of matched pairs and $\text{GIoU}$~\cite{rezatofighi2019generalized} extends IoU with an enclosing-area penalty that provides gradients even when boxes do not overlap. Gradients flow
  through the softmax back into the transformer, encouraging spatially accurate coordinate predictions as part of natural text generation.

  \subsubsection{Curriculum Training.}
  Inspired by curriculum learning~\cite{bengio2009curriculum}, we introduce losses and data modalities gradually over three stages, each building on a stable foundation from the previous one. Figure~\ref{fig:curriculum} illustrates how losses and data modalities are introduced progressively across the three training stages.

  \textit{Stage 1: Classifier warmup.}
  Randomly initialized heads must converge before the base model adapts, otherwise LoRA updates chase noisy gradients from uncalibrated classifiers. Only the classification head is trainable, optimizing
  $\mathcal{L}_\text{bce}$ alone. The training data consists of video-level samples with binary anomaly labels—no text generation targets or spatial annotations are used.

  \textit{Stage 2: Spatial grounding with mixed data.}
  LoRA adapters are unfrozen, enabling the Qwen3-VL backbone to co-adapt with the pre-warmed classification head. This stage introduces two data modalities in an 80-20 mixture:
  (i)~image-level detection samples (80\%), where each input is the last frame of a subclip and the target is a JSON list of object bounding boxes; and (ii)~video-level CoT samples (20\%), where the input is a
  full video and the target is a chain-of-thought response with objects and bounding boxes described inline. The image samples provide dense bounding box supervision, while the video samples teach the model to
  reason over temporal context. The full loss is:
  \begin{equation}
      \mathcal{L}_\text{stage2} = \lambda_\text{bce}\mathcal{L}_\text{bce} + \lambda_\text{lm}\mathcal{L}_\text{lm} + \lambda_\text{giou}\mathcal{L}_\text{giou}.
  \end{equation}
  Learning rate warmup begins from the cosine-decayed~\cite{loshchilov2017sgdr} endpoint of the prior stage to avoid abrupt discontinuities.

  \textit{Stage 3: CoT reasoning.}
  This stage trains exclusively on video-level CoT data, dropping both the image-level detection samples and $\mathcal{L}_\text{giou}$—the model has already internalized spatial
  grounding from Stage~2 and now refines its reasoning quality over full video sequences:
  \begin{equation}
      \mathcal{L}_\text{stage3} = \lambda_\text{bce}\mathcal{L}_\text{bce} + \lambda_\text{lm}\mathcal{L}_\text{lm}.
  \end{equation}

\section{Experiments}
\label{sec:experiments}

\begin{figure*}[t]                                             \centering
      \includegraphics[width=\textwidth]{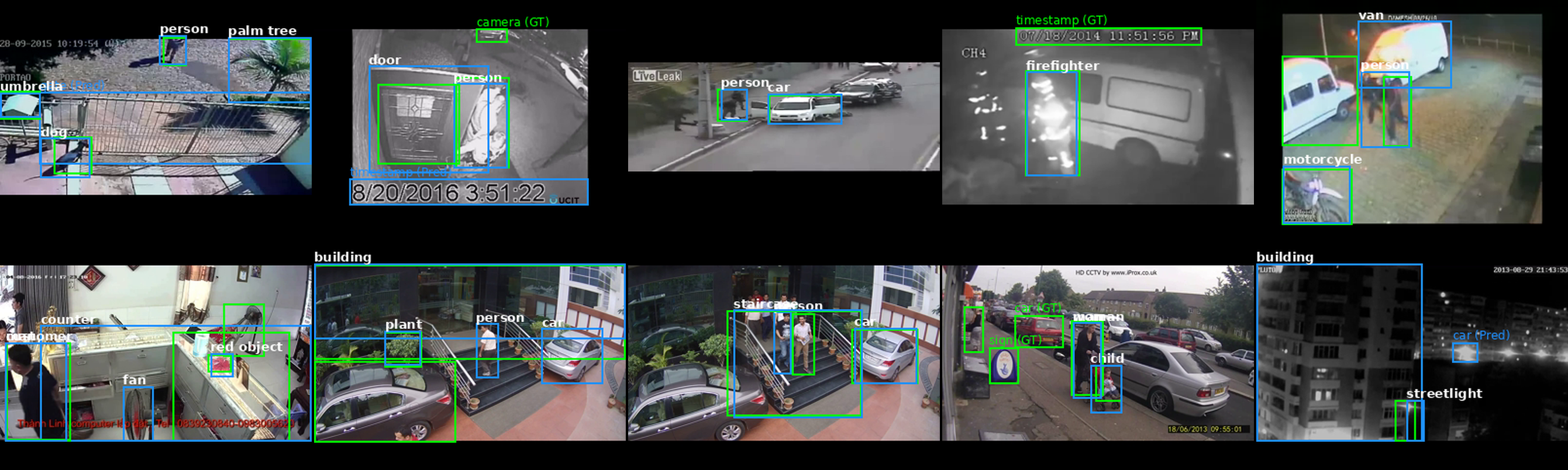}       \caption{Qualitative spatial grounding results on UCF-Crime test samples. Our trained model accurately localizes objects in surveillance frames by predicting bounding boxes (blue) that
   closely align with ground-truth annotations (green). Overlapping labels indicate correctly identified object categories. The model demonstrates robust grounding across diverse scenes     
  including residential areas, streets, fire incidents, and nighttime surveillance.}    \label{fig:spatial_grounding}
  \end{figure*}

\begin{figure*}[t] \centering
      \includegraphics[width=\textwidth]{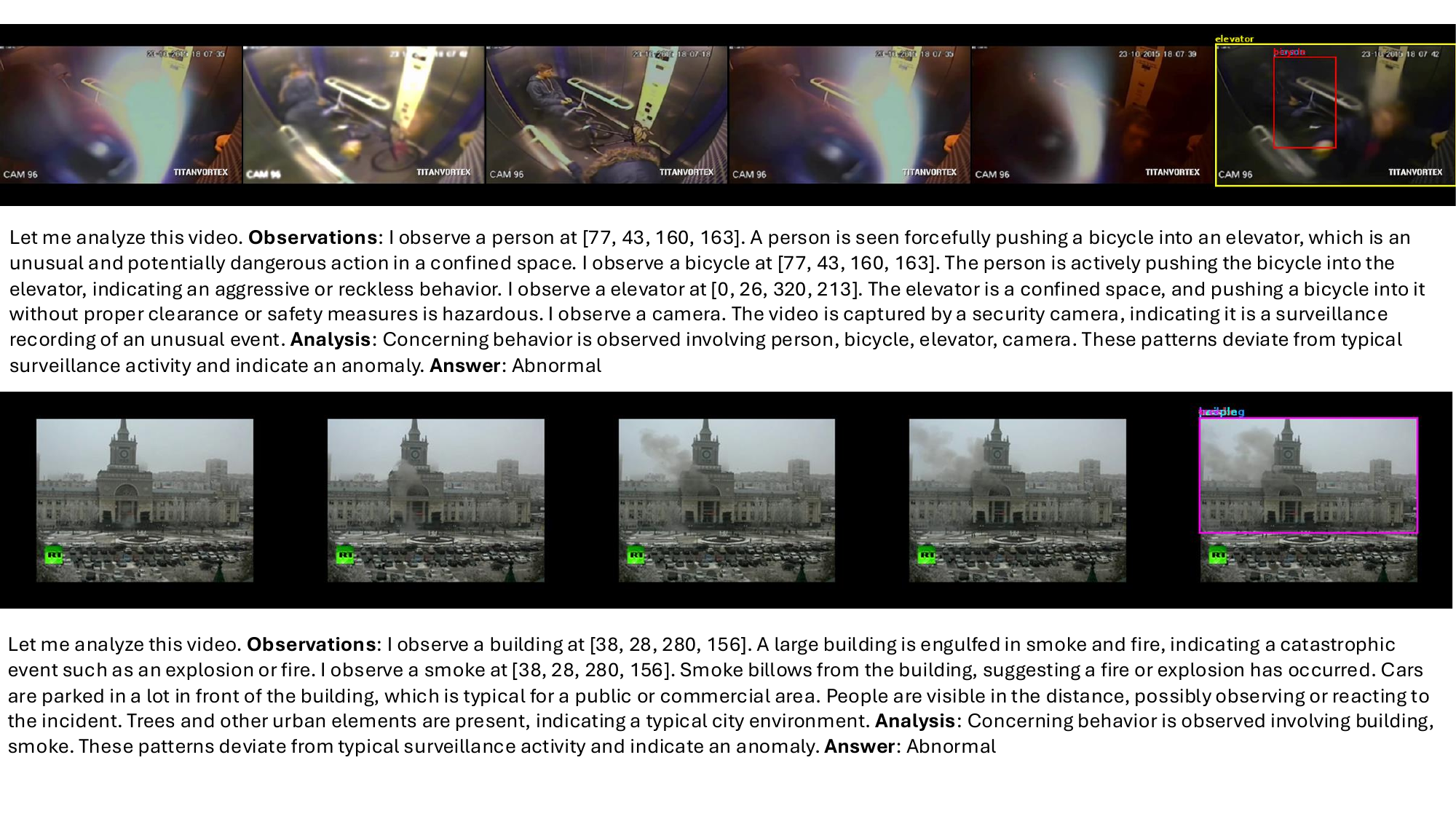} \caption{
      Qualitative CoT-grounded reasoning for video anomaly detection on UCF-Crime test samples. Our model localizes objects in surveillance frames by predicting bounding boxes and  provides per-object reasoning while recording observations. VANGUARD identifies anomalous regions in its analysis and predicts the final answer as "Abnormal" or "Normal." The
   results demonstrate robust explainability and interpretability for video anomaly detection.      
      }    \label{fig:spatial_CoT}
  \end{figure*}

\subsection{Datasets}

We evaluate \textsc{Vanguard} on standard VAD benchmarks, assessing anomaly detection accuracy, reasoning quality, and grounding precision.

\paragraph{Datasets.}
We evaluate on three benchmarks: \textbf{UCF-Crime}~\cite{sultani2018real}, the largest real-world VAD dataset with 1,900 untrimmed surveillance videos (128 hours) across 13 anomaly categories, split into 1,610 training and 290 test videos; \textbf{XD-Violence}~\cite{xdviolence} containing 800 test videos and \textbf{ShanghaiTech Campus}~\cite{liu2018future}, a medium-scale campus surveillance dataset with 107 annotated test videos containing frame-level ground truth annotations. All test videos contain anomalies at varying temporal positions (8.6\%--100\% abnormal frames per video). We evaluate on the available test split. 

\subsection{Implementation Details}

All experiments are conducted on NVIDIA H100 80GB GPUs. We use Qwen3-VL-4B-Instruct~\cite{qwenvl3} as the base VLM, adapted via LoRA~\cite{hu2022lora} ($r=64$, $\alpha=16$, dropout $0.1$) on the query and value projections. We optimize with AdamW~\cite{loshchilov2019decoupled} ($\beta_1=0.9$, $\beta_2=0.999$, $\epsilon=10^{-6}$), cosine learning rate scheduling, weight decay $0.01$, batch size $4$ with gradient accumulation over $2$ steps, and \texttt{bfloat16} mixed precision. The three stages use learning rates of $1e{-}3$, $5e{-}4$, and $2e{-}4$ for $2$, $3$, and $3$ epochs, respectively. We further ablate the contribution of each curriculum stage and the effect of varying the number of subclips per video in the training data; these results are provided in Appendix and confirm that the three-stage curriculum consistently outperforms joint training, and that increasing subclip diversity improves both classification and grounding quality up to a saturation point.

\subsection{Evaluation Metrics}

We evaluate \textsc{Vanguard} along three axes. For anomaly classification, we report ROC-AUC, PR-AUC, Accuracy, Precision, Recall, F1-Score. %and the classifier-text disagreement rate measuring alignment between the classification head and generated text predictions.
For chain-of-thought reasoning quality, we provide qualitative comparisons of generated explanations. For spatial grounding, we report Mean IoU, Recall at IoU threshold of 0.25---the \textbf{first reported spatial grounding metric (bounding-box IoU) for VAD}. % and mean Average Precision (mAP) at the same thresholds %, with predicted boxes extracted from both the bin head (soft argmax over bin logits) and parsed $\langle$bin\_N$\rangle$ tokens in the generated text.

\paragraph{Baselines.}
%We compare against traditional methods (RTFM~\cite{tian2021weakly}, MGFN~\cite{chen2023mgfn}, UR-DMU~\cite{zhou2023urdmu}), zero-shot VLM-based methods (LAVAD~\cite{zanella2024harnessing}, ASK-HINT~\cite{askhint}), and VLM-based fine-tuning methods (Holmes-VAD~\cite{zhang2024holmesvad}, VAD-R1~\cite{huang2025vadr1}). 

We compare against three categories of methods: (1)~\textit{Zero-shot VLM-based methods} that leverage vision-language models without task-specific training, including
  LAVAD~\cite{zanella2024harnessing}, ASK-HINT~\cite{askhint}, VERA~\cite{ye2025vera}, and Flashback~\cite{lee2025flashback}; (2)~\textit{VLM fine-tuning methods} that adapt VLMs for anomaly detection through
  supervised training, including Holmes-VAD~\cite{zhang2024holmesvad} and VAD-R1~\cite{huang2025vadr1}; and (3)~\textit{Grounded VLM methods} that jointly perform spatial localization and classification,
   including Kosmos-2~\cite{peng2023kosmos}, Visual-CoT~\cite{visualcot}, and SpatialVLM~\cite{chen2024spatialvlm}. The first two groups report only VAD classification metrics (AUC, F1), while the third 
  group additionally provides spatial grounding metrics (meanIoU, R@25), enabling direct comparison with \textsc{Vanguard}'s localization capabilities.
  
\subsection{Main Results}

\paragraph{Classification Performance.} 
As shown in Table~\ref{tab:vad_results}, \textsc{Vanguard} achieves 93.78\% AUC and 83.6\% F1 on UCF-Crime, surpassing all zero-shot VLM methods (ASK-HINT 89.83\%,
  Flashback 87.29\%) and fine-tuned baselines (VAD-R1 84.45\%). Even the classifier-warmup stage alone (94.36\% AUC) outperforms all prior methods, confirming that stable head initialization on frozen
  backbone features provides a strong foundation for subsequent fine-tuning. On XD-Violence, \textsc{Vanguard} achieves 91.49\% AUC and 84.03\% F1, competitive with Holmes-VAD (95.88\%) while additionally  
  providing spatial grounding that Holmes-VAD lacks. On ShanghaiTech Campus, frame-level evaluation remains challenging for video-level methods—\textsc{Vanguard} achieves 47.32\% AUC, comparable to VAD-R1
   (51.58\%) and Holmes-VAD (50.00\%), noting that these frame-level metrics ($\star$) are not directly comparable to the video-level protocol used on UCF-Crime and XD-Violence.

\paragraph{Spatial Grounding Quality}
Table~\ref{tab:vad_results} reports spatial grounding metrics alongside detection performance for UCF-Crime test data. Existing VAD methods—both zero-shot (ASK-HINT, VERA, LAVAD, Flashback) and fine-tuned (Holmes-VAD, VAD-R1)—do not provide any spatial grounding, reporting only video-level detection metrics. Among grounded VLM baselines, Kosmos-2 fails entirely at localization (meanIoU 0.00), Visual-CoT achieves moderate overlap (meanIoU 0.45, R@25 0.74) but poor anomaly detection (AUC 0.58), and SpatialVLM offers competitive detection (AUC 0.83) but weaker grounding (meanIoU 0.32, R@25 0.44). As shown in Table~\ref{tab:vad_results}, \textsc{Vanguard} (Spatial CoT) achieves a meanIoU of 0.62 and R@25 of 0.51 while simultaneously attaining the highest detection performance (AUC 0.9378, F1 0.836), demonstrating that interpretable spatial grounding and strong anomaly detection are not mutually exclusive. Qualitatively, Figure~\ref{fig:spatial_grounding} shows how predicted bounding boxes closely align with ground-truth annotations across diverse anomaly types.

It is worth noting that grounded VLM baselines such as Kosmos-2, Visual-CoT, and SpatialVLM were originally trained on large-scale spatial grounding datasets for general-purpose object localization, not for anomaly detection. When applied to surveillance footage, these models can localize common objects but lack the contextual understanding to distinguish normal from anomalous behavior. For instance, SpatialVLM achieves reasonable spatial overlap (meanIoU 0.32) yet its detection AUC of 0.83 lags significantly behind VAD-specific methods. Visual-CoT performs near chance on detection (AUC 0.58) despite achieving the highest R@25 (0.74) among grounded baselines, indicating that localization ability alone does not translate to anomaly understanding. In contrast, \textsc{Vanguard} jointly learns to ground  objects \textit{and} reason about their anomalous context through its chain-of-thought training, bridging the gap between spatial localization and video anomaly detection.

\paragraph{Chain-of-Thought Quality} 
As shown in %Figure~\ref{fig:spatial_CoT}, VANGUARD generates chain-of-thought reasoning grounded in spatial localizations, identifying and describing key objects before analyzing their behavior to arrive at an anomaly prediction. This jointly identifies \emph{what} is anomalous and \emph{where} it appears in the frame.
As shown in Figure~\ref{fig:spatial_grounding}, \textsc{Vanguard} generates structured chain-of-thought reasoning grounded in spatial localizations. The model first records observations by identifying key
  objects and their bounding box coordinates, then provides per-object reasoning describing each object's behavior and its relevance to the scene. In the analysis step, it synthesizes these observations to   
  determine which objects exhibit concerning patterns before arriving at the final anomaly prediction. This structured pipeline jointly identifies \emph{what} is anomalous, \emph{where} it appears in the
  frame, and \emph{why} it constitutes an anomaly—offering a level of interpretability that purely classification-based VAD methods cannot provide.

\paragraph{Limitations and Future Work.}
Visually diffuse or small categories (fire AP@50 0.258, weapon AP@50 0.000) remain challenging due to sparse GroundingDINO annotations, and the pipeline inherits biases from its teacher models. The 4B-parameter VLM backbone also incurs higher inference cost than MIL-based detectors, limiting real-time deployment. Future directions include model distillation, pixel-level segmentation, and faithfulness-aware evaluation of generated explanations.

\paragraph{Impact Statement.}
Interpretable anomaly detection can improve response times, reduce false alarms, and---unlike black-box systems---produce auditable explanations. However, deployment raises privacy concerns and should occur only with appropriate legal authorization. UCF-Crime's predominantly US footage and inherited model biases mean thorough bias auditing is essential. We used only public datasets, designed the system for human oversight, and will release code with clear usage guidelines, urging practitioners to prioritize appropriate safeguards in any surveillance application.

% Conclusion Section
\section{Conclusion}
\label{sec:conclusion}
  We presented \textsc{Vanguard}, a framework that unifies semantic reasoning and spatial grounding for video anomaly detection. Through multi-objective training---from classifier warmup on
  frozen features to spatial-temporal chain-of-thought fine-tuning with text-coordinate GIoU supervision---our method achieves %97.04\% ROC-AUC and 91.76\% F1
  superior performance on UCF-Crime, surpassing prior
  VLM-based approaches. % by over 7 points in AUC.
  Moreover, the model simultaneously produces interpretable spatially grounded bounding boxes (mean IoU 0.619, median
  0.793) along keyframes, with the most frequent categories such as person (IoU 0.873) and vehicle (IoU 0.874) localized with high overlap. Our teacher-student annotation pipeline (Qwen3-VL-4B +
  GroundingDINO) transforms weakly-labeled surveillance data into dense per-keyframe spatial annotations, enabling grounding supervision without manual box labeling.

% ============================================
% REFERENCES
% ============================================
\bibliographystyle{icml2025}
\bibliography{references}

@misc{lee2025flashback,
  title={Flashback: Memory-Driven Zero-shot, Real-time Video Anomaly Detection},
  author={Lee, Hyogun and Kim, Haksub and Kim, Ig-Jae and Choi, Yonghun},
  year={2025},
  eprint={2505.15205},
  archivePrefix={arXiv},
  primaryClass={cs.CV},
  url={https://arxiv.org/abs/2505.15205}
}

@misc{qwenvl3,
      title={Qwen3-VL Technical Report}, 
      author={Shuai Bai and Yuxuan Cai and Ruizhe Chen and Keqin Chen and Xionghui Chen and Zesen Cheng and Lianghao Deng and Wei Ding and Chang Gao and Chunjiang Ge and Wenbin Ge and Zhifang Guo and Qidong Huang and Jie Huang and Fei Huang and Binyuan Hui and Shutong Jiang and Zhaohai Li and Mingsheng Li and Mei Li and Kaixin Li and Zicheng Lin and Junyang Lin and Xuejing Liu and Jiawei Liu and Chenglong Liu and Yang Liu and Dayiheng Liu and Shixuan Liu and Dunjie Lu and Ruilin Luo and Chenxu Lv and Rui Men and Lingchen Meng and Xuancheng Ren and Xingzhang Ren and Sibo Song and Yuchong Sun and Jun Tang and Jianhong Tu and Jianqiang Wan and Peng Wang and Pengfei Wang and Qiuyue Wang and Yuxuan Wang and Tianbao Xie and Yiheng Xu and Haiyang Xu and Jin Xu and Zhibo Yang and Mingkun Yang and Jianxin Yang and An Yang and Bowen Yu and Fei Zhang and Hang Zhang and Xi Zhang and Bo Zheng and Humen Zhong and Jingren Zhou and Fan Zhou and Jing Zhou and Yuanzhi Zhu and Ke Zhu},
      year={2025},
      eprint={2511.21631},
      archivePrefix={arXiv},
      primaryClass={cs.CV},
      url={https://arxiv.org/abs/2511.21631}, 
}

@misc{askhint,
      title={Unlocking Vision-Language Models for Video Anomaly Detection via Fine-Grained Prompting}, 
      author={Shu Zou and Xinyu Tian and Lukas Wesemann and Fabian Waschkowski and Zhaoyuan Yang and Jing Zhang},
      year={2025},
      eprint={2510.02155},
      archivePrefix={arXiv},
      primaryClass={cs.CV},
      url={https://arxiv.org/abs/2510.02155}, 
}

@misc{gpt4v,
      title={GPT-4 Technical Report}, 
      author={OpenAI and Josh Achiam and Steven Adler and Sandhini Agarwal and Lama Ahmad and Ilge Akkaya and Florencia Leoni Aleman and Diogo Almeida and Janko Altenschmidt and Sam Altman and Shyamal Anadkat and Red Avila and Igor Babuschkin and Suchir Balaji and Valerie Balcom and Paul Baltescu and Haiming Bao and Mohammad Bavarian and Jeff Belgum and Irwan Bello and Jake Berdine and Gabriel Bernadett-Shapiro and Christopher Berner and Lenny Bogdonoff and Oleg Boiko and Madelaine Boyd and Anna-Luisa Brakman and Greg Brockman and Tim Brooks and Miles Brundage and Kevin Button and Trevor Cai and Rosie Campbell and Andrew Cann and Brittany Carey and Chelsea Carlson and Rory Carmichael and Brooke Chan and Che Chang and Fotis Chantzis and Derek Chen and Sully Chen and Ruby Chen and Jason Chen and Mark Chen and Ben Chess and Chester Cho and Casey Chu and Hyung Won Chung and Dave Cummings and Jeremiah Currier and Yunxing Dai and Cory Decareaux and Thomas Degry and Noah Deutsch and Damien Deville and Arka Dhar and David Dohan and Steve Dowling and Sheila Dunning and Adrien Ecoffet and Atty Eleti and Tyna Eloundou and David Farhi and Liam Fedus and Niko Felix and Simón Posada Fishman and Juston Forte and Isabella Fulford and Leo Gao and Elie Georges and Christian Gibson and Vik Goel and Tarun Gogineni and Gabriel Goh and Rapha Gontijo-Lopes and Jonathan Gordon and Morgan Grafstein and Scott Gray and Ryan Greene and Joshua Gross and Shixiang Shane Gu and Yufei Guo and Chris Hallacy and Jesse Han and Jeff Harris and Yuchen He and Mike Heaton and Johannes Heidecke and Chris Hesse and Alan Hickey and Wade Hickey and Peter Hoeschele and Brandon Houghton and Kenny Hsu and Shengli Hu and Xin Hu and Joost Huizinga and Shantanu Jain and Shawn Jain and Joanne Jang and Angela Jiang and Roger Jiang and Haozhun Jin and Denny Jin and Shino Jomoto and Billie Jonn and Heewoo Jun and Tomer Kaftan and Łukasz Kaiser and Ali Kamali and Ingmar Kanitscheider and Nitish Shirish Keskar and Tabarak Khan and Logan Kilpatrick and Jong Wook Kim and Christina Kim and Yongjik Kim and Jan Hendrik Kirchner and Jamie Kiros and Matt Knight and Daniel Kokotajlo and Łukasz Kondraciuk and Andrew Kondrich and Aris Konstantinidis and Kyle Kosic and Gretchen Krueger and Vishal Kuo and Michael Lampe and Ikai Lan and Teddy Lee and Jan Leike and Jade Leung and Daniel Levy and Chak Ming Li and Rachel Lim and Molly Lin and Stephanie Lin and Mateusz Litwin and Theresa Lopez and Ryan Lowe and Patricia Lue and Anna Makanju and Kim Malfacini and Sam Manning and Todor Markov and Yaniv Markovski and Bianca Martin and Katie Mayer and Andrew Mayne and Bob McGrew and Scott Mayer McKinney and Christine McLeavey and Paul McMillan and Jake McNeil and David Medina and Aalok Mehta and Jacob Menick and Luke Metz and Andrey Mishchenko and Pamela Mishkin and Vinnie Monaco and Evan Morikawa and Daniel Mossing and Tong Mu and Mira Murati and Oleg Murk and David Mély and Ashvin Nair and Reiichiro Nakano and Rajeev Nayak and Arvind Neelakantan and Richard Ngo and Hyeonwoo Noh and Long Ouyang and Cullen O'Keefe and Jakub Pachocki and Alex Paino and Joe Palermo and Ashley Pantuliano and Giambattista Parascandolo and Joel Parish and Emy Parparita and Alex Passos and Mikhail Pavlov and Andrew Peng and Adam Perelman and Filipe de Avila Belbute Peres and Michael Petrov and Henrique Ponde de Oliveira Pinto and Michael and Pokorny and Michelle Pokrass and Vitchyr H. Pong and Tolly Powell and Alethea Power and Boris Power and Elizabeth Proehl and Raul Puri and Alec Radford and Jack Rae and Aditya Ramesh and Cameron Raymond and Francis Real and Kendra Rimbach and Carl Ross and Bob Rotsted and Henri Roussez and Nick Ryder and Mario Saltarelli and Ted Sanders and Shibani Santurkar and Girish Sastry and Heather Schmidt and David Schnurr and John Schulman and Daniel Selsam and Kyla Sheppard and Toki Sherbakov and Jessica Shieh and Sarah Shoker and Pranav Shyam and Szymon Sidor and Eric Sigler and Maddie Simens and Jordan Sitkin and Katarina Slama and Ian Sohl and Benjamin Sokolowsky and Yang Song and Natalie Staudacher and Felipe Petroski Such and Natalie Summers and Ilya Sutskever and Jie Tang and Nikolas Tezak and Madeleine B. Thompson and Phil Tillet and Amin Tootoonchian and Elizabeth Tseng and Preston Tuggle and Nick Turley and Jerry Tworek and Juan Felipe Cerón Uribe and Andrea Vallone and Arun Vijayvergiya and Chelsea Voss and Carroll Wainwright and Justin Jay Wang and Alvin Wang and Ben Wang and Jonathan Ward and Jason Wei and CJ Weinmann and Akila Welihinda and Peter Welinder and Jiayi Weng and Lilian Weng and Matt Wiethoff and Dave Willner and Clemens Winter and Samuel Wolrich and Hannah Wong and Lauren Workman and Sherwin Wu and Jeff Wu and Michael Wu and Kai Xiao and Tao Xu and Sarah Yoo and Kevin Yu and Qiming Yuan and Wojciech Zaremba and Rowan Zellers and Chong Zhang and Marvin Zhang and Shengjia Zhao and Tianhao Zheng and Juntang Zhuang and William Zhuk and Barret Zoph},
      year={2023},
      eprint={2303.08774},
      archivePrefix={arXiv},
      primaryClass={cs.CL},
      url={https://arxiv.org/abs/2303.08774},
}

@inproceedings{sultani2018real,
  title={Real-world anomaly detection in surveillance videos},
  author={Sultani, Waqas and Chen, Chen and Shah, Mubarak},
  booktitle={Proceedings of the IEEE conference on computer vision and pattern recognition},
  pages={6479--6488},
  year={2018}
}

@article{Zhou_2022,
   title={Learning to Prompt for Vision-Language Models},
   volume={130},
   ISSN={1573-1405},
   url={http://dx.doi.org/10.1007/s11263-022-01653-1},
   DOI={10.1007/s11263-022-01653-1},
   number={9},
   journal={International Journal of Computer Vision},
   publisher={Springer Science and Business Media LLC},
   author={Zhou, Kaiyang and Yang, Jingkang and Loy, Chen Change and Liu, Ziwei},
   year={2022},
   month=jul, pages={2337–2348} }

@misc{pu2024learningpromptenhancedcontextfeatures,
      title={Learning Prompt-Enhanced Context Features for Weakly-Supervised Video Anomaly Detection}, 
      author={Yujiang Pu and Xiaoyu Wu and Lulu Yang and Shengjin Wang},
      year={2024},
      eprint={2306.14451},
      archivePrefix={arXiv},
      primaryClass={cs.CV},
      url={https://arxiv.org/abs/2306.14451}, 
}

@article{Kim_Yoon_Choi_Sull_2023,
  title={Unsupervised Video Anomaly Detection Based on Similarity with Predefined Text Descriptions},
  author={Kim, Jaehyun and Yoon, Seongwook and Choi, Taehyeon and Sull, Sanghoon},
  journal={Sensors},
  volume={23},
  number={14},
  pages={6256},
  year={2023},
  publisher={MDPI},
  doi={10.3390/s23146256},
  url={https://www.mdpi.com/1424-8220/23/14/6256}
}

@misc{visualcot,
      title={Visual CoT: Advancing Multi-Modal Language Models with a Comprehensive Dataset and Benchmark for Chain-of-Thought Reasoning}, 
      author={Hao Shao and Shengju Qian and Han Xiao and Guanglu Song and Zhuofan Zong and Letian Wang and Yu Liu and Hongsheng Li},
      year={2024},
      eprint={2403.16999},
      archivePrefix={arXiv},
      primaryClass={cs.CV},
      url={https://arxiv.org/abs/2403.16999}, 
}

@misc{ucadataset,
      title={Towards Surveillance Video-and-Language Understanding: New Dataset, Baselines, and Challenges}, 
      author={Tongtong Yuan and Xuange Zhang and Kun Liu and Bo Liu and Chen Chen and Jian Jin and Zhenzhen Jiao},
      year={2023},
      eprint={2309.13925},
      archivePrefix={arXiv},
      primaryClass={cs.CV}
}

@inproceedings{wu2023vadclip,
  title={{VadCLIP}: Adapting vision-language models for weakly supervised video anomaly detection},
  author={Wu, Peng and Zhou, Xuerong and Pang, Guansong and Zhou, Lingru and Yan, Qingsen and Wang, Peng and Zhang, Yanning},
  booktitle={Proceedings of the AAAI Conference on Artificial Intelligence},
  volume={38},
  pages={6074--6082},
  year={2024}
}

@inproceedings{chen2022pix2seq,                                                                                                                                                     
    title={Pix2Seq: A Language Modeling Framework for Object Detection},                                                                                                              
    author={Chen, Ting and Saxena, Saurabh and Li, Lala and Fleet, David J and Hinton, Geoffrey},                                                                                     
    booktitle={International Conference on Learning Representations (ICLR)},                                                                                                          
    year={2022}                                                                                                                                                                       
  }

@article{nibali2018numerical,                                                                                                                                                       
    title={Numerical Coordinate Regression with Convolutional Neural Networks},
    author={Nibali, Aiden and He, Zhen and Morgan, Stuart and Prendergast, Luke},
    journal={arXiv preprint arXiv:1801.07372},
    year={2018}
  }

@inproceedings{rezatofighi2019generalized,
    title={Generalized Intersection Over Union: A Metric and a Loss for Bounding Box Regression},
    author={Rezatofighi, Hamid and Tsoi, Nathan and Gwak, JunYoung and Sadeghian, Amir and Reid, Ian and Savarese, Silvio},
    booktitle={Proceedings of the IEEE/CVF Conference on Computer Vision and Pattern Recognition (CVPR)},
    pages={658--666},
    year={2019}
  }

@inproceedings{radford2021clip,                                   
    title     = {Learning Transferable Visual Models From Natural Language Supervision},                                                                                                            
    author    = {Radford, Alec and Kim, Jong Wook and Hallacy, Chris and Ramesh, Aditya and Goh, Gabriel and Agarwal, Sandhini and Sastry, Girish and Askell, Amanda and Mishkin, Pamela and Clark, 
  Jack and Krueger, Gretchen and Sutskever, Ilya},
    booktitle = {Proceedings of the 38th International Conference on Machine Learning (ICML)},                                                                                                      
    pages     = {8748--8763},
    year      = {2021},
    publisher = {PMLR}
  }

@article{Li_Liu_Jiao_2022,
  title={Self-Training Multi-Sequence Learning with Transformer for Weakly Supervised Video Anomaly Detection},
  volume={36},
  url={https://ojs.aaai.org/index.php/AAAI/article/view/20028},
  DOI={10.1609/aaai.v36i2.20028},
  abstractNote={Weakly supervised Video Anomaly Detection (VAD) using Multi-Instance Learning (MIL) is usually based on the fact that the anomaly score of an abnormal snippet is higher than that of a normal snippet. In the beginning of training, due to the limited accuracy of the model, it is easy to select the wrong abnormal snippet. In order to reduce the probability of selection errors, we first propose a Multi-Sequence Learning (MSL) method and a hinge-based MSL ranking loss that uses a sequence composed of multiple snippets as an optimization unit. We then design a Transformer-based MSL network to learn both video-level anomaly probability and snippet-level anomaly scores. In the inference stage, we propose to use the video-level anomaly probability to suppress the fluctuation of snippet-level anomaly scores. Finally, since VAD needs to predict the snippet-level anomaly scores, by gradually reducing the length of selected sequence, we propose a self-training strategy to gradually refine the anomaly scores. Experimental results show that our method achieves significant improvements on ShanghaiTech, UCF-Crime, and XD-Violence.},
  number={2},
  journal={Proceedings of the AAAI Conference on Artificial Intelligence},
  author={Li, Shuo and Liu, Fang and Jiao, Licheng},
  year={2022},
  month={Jun.},
  pages={1395-1403}
}

@InProceedings{pmlr-v267-huang25ad,
  title = 	 {Ex-{VAD}: Explainable Fine-grained Video Anomaly Detection Based on Visual-Language Models},
  author =       {Huang, Chao and Shi, Yushu and Wen, Jie and Wang, Wei and Xu, Yong and Cao, Xiaochun},
  booktitle = 	 {Proceedings of the 42nd International Conference on Machine Learning},
  pages = 	 {25750--25761},
  year = 	 {2025},
  editor = 	 {Singh, Aarti and Fazel, Maryam and Hsu, Daniel and Lacoste-Julien, Simon and Berkenkamp, Felix and Maharaj, Tegan and Wagstaff, Kiri and Zhu, Jerry},
  volume = 	 {267},
  series = 	 {Proceedings of Machine Learning Research},
  month = 	 {13--19 Jul},
  publisher =    {PMLR},
  url = 	 {https://proceedings.mlr.press/v267/huang25ad.html}
}

@inproceedings{tian2021weakly,
  title={Weakly-supervised video anomaly detection with robust temporal feature magnitude learning},
  author={Tian, Yu and Pang, Guansong and Chen, Yuanhong and Singh, Rajvinder and Verjans, Johan W and Carneiro, Gustavo},
  booktitle={Proceedings of the IEEE/CVF International Conference on Computer Vision},
  pages={4975--4986},
  year={2021}
}

@inproceedings{chen2023mgfn,
  title={MGFN: Magnitude-contrastive glance-and-focus network for weakly-supervised video anomaly detection},
  author={Chen, Yingxian and Liu, Zhengzhe and Zhang, Baoheng and Fok, Wilton and Qi, Xiaojuan and Wu, Yik-Chung},
  booktitle={Proceedings of the AAAI Conference on Artificial Intelligence},
  volume={37},
  pages={387--395},
  year={2023}
}

@inproceedings{zhou2023urdmu,
  title={Dual Memory Units with Uncertainty Regulation for Weakly Supervised Video Anomaly Detection},
  author={Zhou, Hang and Yu, Junqing and Yang, Wei},
  booktitle={Proceedings of the AAAI Conference on Artificial Intelligence},
  volume={37},
  pages={3769--3777},
  year={2023}
}

@inproceedings{liu2018future,
  title={Future frame prediction for anomaly detection--a new baseline},
  author={Liu, Wen and Luo, Weixin and Lian, Dongze and Gao, Shenghua},
  booktitle={Proceedings of the IEEE Conference on Computer Vision and Pattern Recognition},
  pages={6536--6545},
  year={2018}
}

@inproceedings{hasan2016learning,
  title={Learning temporal regularity in video sequences},
  author={Hasan, Mahmudul and Choi, Jonghyun and Neumann, Jan and Roy-Chowdhury, Amit K and Davis, Larry S},
  booktitle={Proceedings of the IEEE Conference on Computer Vision and Pattern Recognition},
  pages={733--742},
  year={2016}
}

@inproceedings{park2020learning,
  title={Learning memory-guided normality for anomaly detection},
  author={Park, Hyunjong and Noh, Jongyoun and Ham, Bumsub},
  booktitle={Proceedings of the IEEE/CVF Conference on Computer Vision and Pattern Recognition},
  pages={14372--14381},
  year={2020}
}

@inproceedings{gong2019memorizing,
  title={Memorizing normality to detect anomaly: Memory-augmented deep autoencoder for unsupervised anomaly detection},
  author={Gong, Dong and Liu, Lingqiao and Le, Vuong and Saha, Budhaditya and Mansour, Moussa Reda and Venkatesh, Svetha and Hengel, Anton van den},
  booktitle={Proceedings of the IEEE/CVF International Conference on Computer Vision},
  pages={1705--1714},
  year={2019}
}

@inproceedings{zanella2024harnessing,
  title={Harnessing large language models for training-free video anomaly detection},
  author={Zanella, Luca and Menapace, Willi and Mancini, Massimiliano and Wang, Yiming and Ricci, Elisa},
  booktitle={Proceedings of the IEEE/CVF Conference on Computer Vision and Pattern Recognition},
  pages={18527--18536},
  year={2024}
}

@inproceedings{ye2025vera,
  title={VERA: Explainable video anomaly detection via verbalized learning of vision-language models},
  author={Ye, Muchao and Liu, Weiyang and He, Pan},
  booktitle={Proceedings of the IEEE/CVF Conference on Computer Vision and Pattern Recognition},
  year={2025}
}

@inproceedings{li2024vlavad,
  title={{VLAVAD}: Vision-Language Models Assisted Unsupervised Video Anomaly Detection},
  author={Li, Changkang and Jiang, Yalong},
  booktitle={Proceedings of the British Machine Vision Conference},
  year={2024}
}

@article{zhang2024holmesvad,
  title={{Holmes-VAD}: Towards unbiased and explainable video anomaly detection via multi-modal large language model},
  author={Zhang, Huaxin and others},
  journal={arXiv preprint arXiv:2406.12235},
  year={2024}
}

@article{bai2023qwen,
  title={Qwen-VL: A versatile vision-language model for understanding, localization, text reading, and beyond},
  author={Bai, Jinze and Bai, Shuai and Yang, Shusheng and Wang, Shijie and Tan, Sinan and Wang, Peng and Lin, Junyang and Zhou, Chang and Zhou, Jingren},
  journal={arXiv preprint arXiv:2308.12966},
  year={2023}
}

@InProceedings{ucf_crime,
author = {Sultani, Waqas and Chen, Chen and Shah, Mubarak},
title = {Real-World Anomaly Detection in Surveillance Videos},
booktitle = {The IEEE Conference on Computer Vision and Pattern Recognition (CVPR)},
month = {June},
year = {2018}
}

@InProceedings{Lu_2013_ICCV,
author = {Lu, Cewu and Shi, Jianping and Jia, Jiaya},
title = {Abnormal Event Detection at 150 FPS in MATLAB},
booktitle = {Proceedings of the IEEE International Conference on Computer Vision (ICCV)},
month = {December},
year = {2013}
}

@InProceedings{Luo_2017_ICCV,
author = {Luo, Weixin and Liu, Wen and Gao, Shenghua},
title = {A Revisit of Sparse Coding Based Anomaly Detection in Stacked RNN Framework},
booktitle = {Proceedings of the IEEE International Conference on Computer Vision (ICCV)},
month = {Oct},
year = {2017}
}

@article{Wang2010AnomalyDI,
  title={Anomaly detection in crowd scene},
  author={Shu Wang and Zhenjiang Miao},
  journal={IEEE 10th INTERNATIONAL CONFERENCE ON SIGNAL PROCESSING PROCEEDINGS},
  year={2010},
  pages={1220-1223},
  url={https://api.semanticscholar.org/CorpusID:2742268}
}

@inproceedings{xdviolence,
title={Not only Look, but also Listen: Learning Multimodal Violence Detection under 
Weak Supervision},
author={Wu, Peng and Liu, jing and Shi, Yujia and Sun, Yujia and Shao, Fangtao 
and Wu, Zhaoyang and Yang, Zhiwei},
booktitle={European Conference on Computer Vision (ECCV)},
year={2020}
}

@inproceedings{liu2024visual,
  title={Visual instruction tuning},
  author={Liu, Haotian and Li, Chunyuan and Wu, Qingyang and Lee, Yong Jae},
  booktitle={Advances in Neural Information Processing Systems},
  volume={36},
  year={2023}
}

@inproceedings{liu2024grounding,
  title={Grounding {DINO}: Marrying {DINO} with grounded pre-training for open-set object detection},
  author={Liu, Shilong and Zeng, Zhaoyang and Ren, Tianhe and Li, Feng and Zhang, Hao and Yang, Jie and Li, Chunyuan and Yang, Jianwei and Su, Hang and Zhu, Jun and others},
  booktitle={European Conference on Computer Vision},
  pages={38--55},
  year={2024},
  organization={Springer}
}

@misc{bai2025qwen25vltechnicalreport,
      title={Qwen2.5-VL Technical Report}, 
      author={Shuai Bai and Keqin Chen and Xuejing Liu and Jialin Wang and Wenbin Ge and Sibo Song and Kai Dang and Peng Wang and Shijie Wang and Jun Tang and Humen Zhong and Yuanzhi Zhu and Mingkun Yang and Zhaohai Li and Jianqiang Wan and Pengfei Wang and Wei Ding and Zheren Fu and Yiheng Xu and Jiabo Ye and Xi Zhang and Tianbao Xie and Zesen Cheng and Hang Zhang and Zhibo Yang and Haiyang Xu and Junyang Lin},
      year={2025},
      eprint={2502.13923},
      archivePrefix={arXiv},
      primaryClass={cs.CV},
      url={https://arxiv.org/abs/2502.13923}, 
}

@inproceedings{rasheed2024glamm,
  title={{GLaMM}: Pixel grounding large multimodal model},
  author={Rasheed, Hanoona and Maaz, Muhammad and Shaker, Abdelrahman and Khan, Salman and Cholakal, Hisham and Anwer, Rao Muhammad and Xia, Gui-Song and Shi, Dianbing and Khan, Fahad Shahbaz},
  booktitle={Proceedings of the IEEE/CVF Conference on Computer Vision and Pattern Recognition},
  pages={13009--13018},
  year={2024}
}

@inproceedings{peng2023kosmos,
  title={Kosmos-2: Grounding multimodal large language models to the world},
  author={Peng, Zhiliang and Wang, Wenhui and Dong, Li and Hao, Yaru and Huang, Shaohan and Ma, Shuming and Wei, Furu},
  booktitle={International Conference on Learning Representations},
  year={2024}
}

@misc{chen2023shikra,
  title={Shikra: Unleashing multimodal {LLM's} referential dialogue magic},
  author={Chen, Keqin and Zhang, Zhao and Zeng, Weili and Zhang, Richong and Zhu, Feng and Zhao, Rui},
  year={2023},
  eprint={2306.15195},
  archivePrefix={arXiv},
  primaryClass={cs.CV}
}

@inproceedings{chen2024spatialvlm,
  title={{SpatialVLM}: Endowing vision-language models with spatial reasoning capabilities},
  author={Chen, Boyuan and Xu, Zhuo and Kirmani, Sean and Ichter, Brian and Driess, Danny and Florence, Pete and Sadigh, Dorsa and Guibas, Leonidas and Xia, Fei},
  booktitle={Proceedings of the IEEE/CVF Conference on Computer Vision and Pattern Recognition},
  pages={14455--14465},
  year={2024}
}

@InProceedings{Bhowmik_2025_ICCV,
    author    = {Bhowmik, Aritra and Derakhshani, Mohammad Mahdi and Koelma, Dennis and Asano, Yuki M. and Oswald, Martin R. and Snoek, Cees G. M.},
    title     = {TWIST \& SCOUT: Grounding Multimodal LLM-Experts by Forget-Free Tuning},
    booktitle = {Proceedings of the IEEE/CVF International Conference on Computer Vision (ICCV)},
    month     = {October},
    year      = {2025},
    pages     = {1359-1368}
}

@inproceedings{yang2024anomalyruler,
  title={{AnomalyRuler}: Follow the rules: Reasoning for video anomaly detection with large language models},
  author={Yang, Yuchen and others},
  booktitle={European Conference on Computer Vision},
  pages={304--322},
  year={2024},
  organization={Springer}
}

@inproceedings{tang2024hawk,
  title={{HAWK}: Learning to understand open-world video anomalies},
  author={Tang, Jiaqi and Lu, Hao and Wu, Ruizheng and Xu, Xiaogang and Ma, Ke and Fang, Cheng and Guo, Bin and Lu, Jiangbo and Chen, Qifeng and Chen, Ying-Cong},
  booktitle={Advances in Neural Information Processing Systems},
  volume={37},
  year={2024}
}

@article{huang2025vadr1,
  title={{Vad-R1}: Towards video anomaly reasoning via perception-to-cognition chain-of-thought},
  author={Huang, Chao and Wang, Benfeng and others},
  journal={arXiv preprint arXiv:2505.19877},
  year={2025},
  note={Accepted at NeurIPS 2025}
}

@article{vagu2025,
  title={{VAGU \& GtS}: {LLM}-based benchmark and framework for joint video anomaly grounding and understanding},
  author={Gao, Shibo and Yang, Peipei and Liu, Yangyang and Chen, Yi and Zhu, Han and Zhang, Xuyao and Huang, Linlin},
  journal={arXiv preprint arXiv:2507.21507},
  year={2025},
  note={Accepted at BMVC 2025}
}

@inproceedings{wei2022chain,
  title={Chain-of-thought prompting elicits reasoning in large language models},
  author={Wei, Jason and Wang, Xuezhi and Schuurmans, Dale and Bosma, Maarten and Xia, Fei and Chi, Ed and Le, Quoc V and Zhou, Denny and others},
  booktitle={Advances in Neural Information Processing Systems},
  volume={35},
  pages={24824--24837},
  year={2022}
}

@inproceedings{carion2020end,
  title={End-to-end object detection with transformers},
  author={Carion, Nicolas and Massa, Francisco and Synnaeve, Gabriel and Usunier, Nicolas and Kirillov, Alexander and Zagoruyko, Sergey},
  booktitle={European Conference on Computer Vision},
  pages={213--229},
  year={2020},
  organization={Springer}
}

@inproceedings{hu2022lora,
  title={{LoRA}: Low-Rank Adaptation of Large Language Models},
  author={Hu, Edward J and Shen, Yelong and Wallis, Phillip and Allen-Zhu, Zeyuan and Li, Yuanzhi and Wang, Shean and Wang, Lu and Chen, Weizhu},
  booktitle={International Conference on Learning Representations},
  year={2022},
  url={https://openreview.net/forum?id=nZeVKeeFYf9}
}

@inproceedings{loshchilov2019decoupled,
  title={Decoupled Weight Decay Regularization},
  author={Loshchilov, Ilya and Hutter, Frank},
  booktitle={International Conference on Learning Representations},
  year={2019},
  url={https://openreview.net/forum?id=Bkg6RiCqY7}
}

@inproceedings{bengio2009curriculum,
  author    = {Bengio, Yoshua and Louradour, J{\'{e}}r{\^{o}}me and Collobert, Ronan and Weston, Jason},
  title     = {Curriculum Learning},
  booktitle = {Proceedings of the 26th Annual International Conference on Machine Learning},
  pages     = {41--48},
  year      = {2009},
  publisher = {ACM},
  doi       = {10.1145/1553374.1553380}
}

@inproceedings{sener2018multitask,
  author    = {Sener, Ozan and Koltun, Vladlen},
  title     = {Multi-Task Learning as Multi-Objective Optimization},
  booktitle = {Advances in Neural Information Processing Systems},
  volume    = {31},
  year      = {2018}
}

@inproceedings{kendall2018multitask,
  title     = {Multi-Task Learning Using Uncertainty to Weigh Losses for Scene Geometry and Semantics},
  author    = {Kendall, Alex and Gal, Yarin and Cipolla, Roberto},
  booktitle = {Proceedings of the IEEE/CVF Conference on Computer Vision and Pattern Recognition},
  pages     = {7482--7491},
  year      = {2018}
}

@inproceedings{loshchilov2017sgdr,
  author    = {Loshchilov, Ilya and Hutter, Frank},
  title     = {{SGDR}: Stochastic Gradient Descent with Warm Restarts},
  booktitle = {International Conference on Learning Representations},
  year      = {2017}
}

@inproceedings{sun2018integral,
  title     = {Integral Human Pose Regression},
  author    = {Sun, Xiao and Xiao, Bin and Wei, Fangyin and Liang, Shuang and Wei, Yichen},
  booktitle = {Proceedings of the European Conference on Computer Vision},
  pages     = {536--553},
  year      = {2018},
  publisher = {Springer}
}

@inproceedings{wei2022finetuned,
  author    = {Wei, Jason and Bosma, Maarten and Zhao, Vincent Y. and Guu, Kelvin and Yu, Adams Wei and Lester, Brian and Du, Nan and Dai, Andrew M. and Le, Quoc V.},
  title     = {Finetuned Language Models Are Zero-Shot Learners},
  booktitle = {International Conference on Learning Representations},
  year      = {2022}
}

@inproceedings{wang2023selfinstruct,
  title     = {Self-Instruct: Aligning Language Models with Self-Generated Instructions},
  author    = {Wang, Yizhong and Kordi, Yeganeh and Mishra, Swaroop and Liu, Alisa and Smith, Noah A. and Khashabi, Daniel and Hajishirzi, Hannaneh},
  booktitle = {Proceedings of the 61st Annual Meeting of the Association for Computational Linguistics},
  pages     = {13484--13508},
  year      = {2023}
}

% ============================================
% SUPPLEMENTARY MATERIAL
% ============================================
\clearpage
\appendix
% Supplementary Material for VANGUARD
% Video Anomaly uNderstandinG throUgh reAsoning and gRounDing

\section*{Supplementary Material}
This supplementary document provides additional details and analyses that complement the main paper. Section~\ref{sec:supp_baselines} describes each baseline method used in our comparisons. Section~\ref{sec:supp_qualitative} presents qualitative side-by-side comparisons of spatial grounding and chain-of-thought reasoning between baselines and \textsc{Vanguard}. Section~\ref{sec:supp_benchmark} elaborates on the VANGUARD-Bench construction pipeline, including the exact prompts, input/output formats, dataset statistics, and object category taxonomy. Section~\ref{sec:supp_implementation} provides full implementation details covering model architecture, per-stage training hyperparameters, loss weight ($\lambda$) configurations with a sensitivity analysis, and computational cost. Section~\ref{sec:supp_ablations} presents ablation studies on curriculum training versus joint optimization and the effect of varying the number of training subclips per video. Section~\ref{sec:supp_inference} describes real-time inference deployment using event-driven CLIP-based scene gating. Finally, Section~\ref{sec:supp_biases} discusses biases inherited from the VANGUARD-Bench construction pipeline, including subclipping strategy, last-frame anchoring, teacher model cascading errors, and geographic skew.

% =====================================================================
\section{Baseline Descriptions}
\label{sec:supp_baselines}

We compare \textsc{Vanguard} against three categories of methods spanning zero-shot inference, VLM fine-tuning, and grounded spatial reasoning. Below we describe each baseline and how it was evaluated.

\subsection{Zero-Shot VLM-Based Methods}
For ASK-HINT~\cite{askhint}, VERA~\cite{ye2025vera}, LaVAD~\cite{li2024vlavad}, and Flashback~\cite{lee2025flashback}, we report AUC scores directly from their respective papers, as these methods use standardized evaluation protocols on the same UCF-Crime and XD-Violence test splits. No re-implementation or re-evaluation was performed for these baselines.

\subsection{VLM Fine-Tuning Methods}

\paragraph{Holmes-VAD~\cite{zhang2024holmesvad}.}
 We use the publicly available HolmesVAD-7B. %checkpoint (\texttt{ppxin321/HolmesVAD-7B}), built on VideoLLaVA with a LanguageBind video tower and Vicuna-7B + LoRA.
 Short clips ($<$200 frames) are sampled at 8 frames; long videos are split into 5 non-overlapping temporal windows with independent inference. The model is prompted as follows:

  \begin{quote}\small\ttfamily
  Are there any abnormal events in this video clip?
  \end{quote}

  The generated response is parsed with a priority-ordered heuristic: explicit keywords (\textit{``yes''}/\textit{``abnormal''} $\to$ anomalous; \textit{``no''}/\textit{``normal''} $\to$ normal), with fallback matching for \textit{``anomal''}, \textit{``unusual''}, and \textit{``suspicious''} (excluding negated forms). The video-level score for long videos is the fraction of windows predicting abnormal. Inference uses FP16, greedy decoding (temperature $0$, max 128 tokens).

  \paragraph{VAD-R1~\cite{huang2025vadr1}.}
  We evaluate the publicly available VAD-R1 checkpoint. We sample 16 frames per video (max 100{,}352 pixels each). The model receives:

  \begin{quote}\small
  \textbf{System:} \texttt{You are a multimodal reasoning assistant for understanding anomalies in videos.}

  \textbf{User:} Generate a \texttt{<think>} block with a 4-step reasoning chain: (1)~Scene Description, (2)~Abnormal Event Description with spatial location, (3)~Abnormal Event Recognition, (4)~Causal Reasoning and Social Norms. Then output an \texttt{<answer>} block with XML tags: \texttt{<which>} (Normal/Abnormal), \texttt{<what>} (description), \texttt{<when>} (temporal interval, e.g., $[0.25, 0.45]$), \texttt{<where>} (spatial description), \texttt{<why>} (justification).
  \end{quote}

  Binary classification is extracted from \texttt{<which>}, temporal localization from \texttt{<when>}, and frame-level scores are generated by converting predicted intervals to per-frame values with Gaussian smoothing. Inference uses vLLM with bfloat16 (temperature 0.1, top-p 0.9, max 768 tokens).

\begin{figure*}[!htbp]
    \centering
    \includegraphics[width=\textwidth]{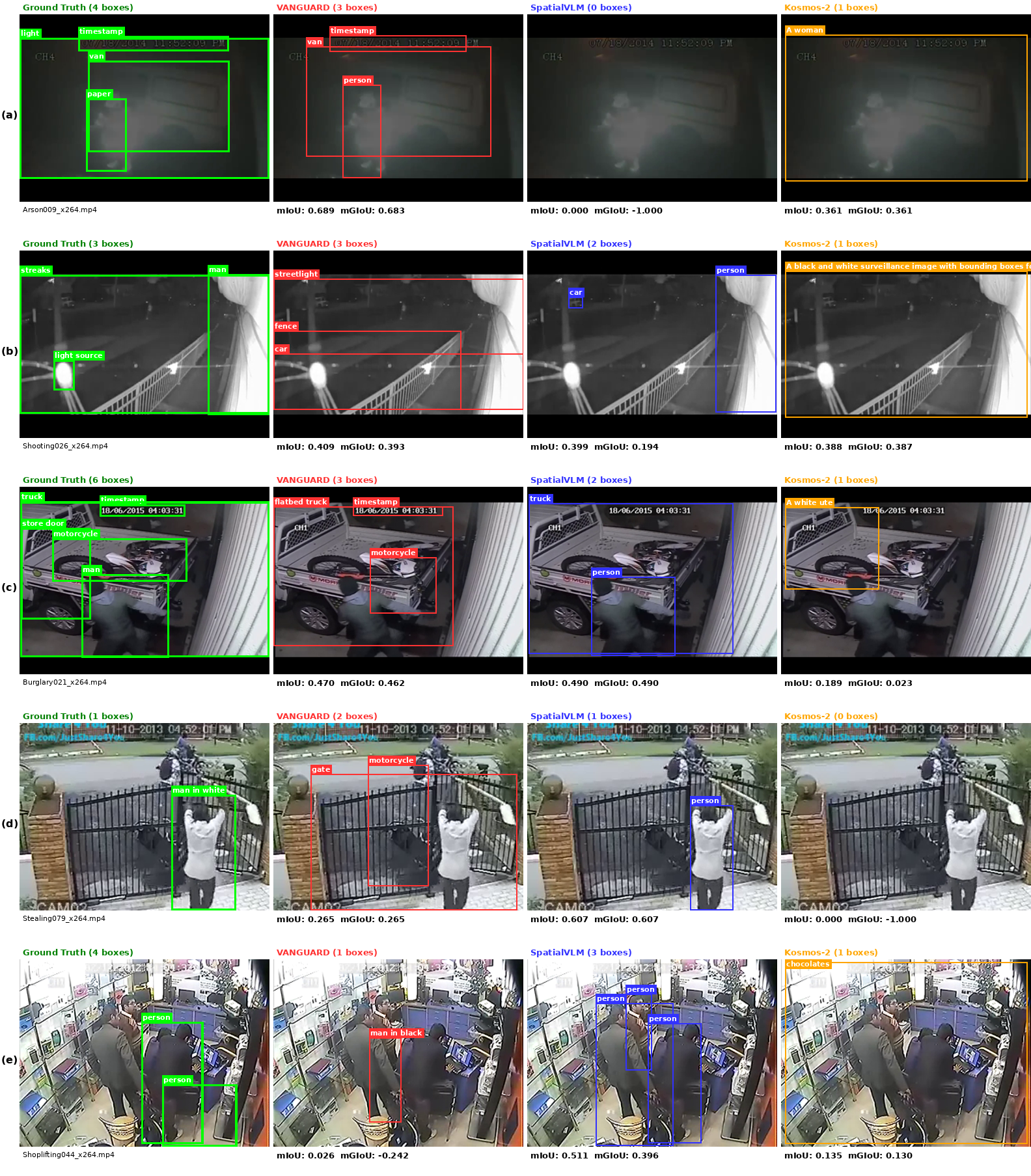}
    \caption{Qualitative comparison of spatial grounding across methods on UCF-Crime test samples. Each row shows a different test video frame. Columns from left to right:
  ground-truth annotations (green boxes with object labels), SpatialVLM, Kosmos-2, and \textsc{Vanguard} (red boxes). \textsc{Vanguard} produces tighter, more
  accurate bounding boxes with correct object-level anomaly attribution---localizing anomalous entities and their context (e.g., fire and van in Arson009 (a),
  motorcycle on the ground in Stealing079 (d))---while baseline methods either output single coarse bounding boxes covering the entire scene (Kosmos-2), or detect
  generic foreground objects irrespective of anomaly relevance (SpatialVLM). Notably in Shoplifting044 (e), SpatialVLM localizes bystanders rather than the
  shoplifting act. \textsc{Vanguard} also demonstrates stronger robustness in low-visibility night scenes such as Arson009 (a) and Shooting026 (b), where both
  baselines either fail to produce any boxes or yield imprecise localizations.}
    \label{fig:supp_spatial_comparison}
\end{figure*}

  \subsection{Grounded VLM Methods}

  These baselines produce spatial bounding boxes alongside predictions, enabling direct comparison with \textsc{Vanguard}'s localization capabilities.

  \paragraph{Kosmos-2~\cite{peng2023kosmos}.}
  We use \texttt{microsoft/kosmos-2-patch14-224} in VQA text-generation mode (without the \texttt{<grounding>} prefix). We sample 8 frames for short clips ($<$200 frames) or 16 for long videos, processing each frame independently. The prompt is:

  \begin{quote}\small\ttfamily
  Question: This is a frame from a surveillance camera. Is there any crime, violence, fight, weapon, accident, or abnormal activity visible? Answer with `Abnormal' if something suspicious or dangerous is happening, or `Normal' if the scene looks ordinary. Answer:
  \end{quote}

  The response is parsed with priority on \textit{``Abnormal''}/\textit{``Normal''} keywords in the first 80 characters, with fallback matching for violence-related terms. The video-level score is the fraction of frames classified as abnormal (threshold 0.5). Inference uses FP16, greedy decoding (max 64 tokens).

  \paragraph{Visual-CoT~\cite{visualcot}.}
  We evaluate \texttt{VisCoT-7b-336} (\texttt{deepcs233/VisCoT-7b-336}), a LLaVA-based model fine-tuned for visual CoT with spatial localization. Inference follows a two-turn protocol (8 frames for short clips, 16 for long videos):

  \begin{quote}\small
  \textbf{Turn 1:} \texttt{You are a highly advanced Video Surveillance System. Analyze this surveillance camera frame. Consider: (1) people in unusual positions or non-typical behaviour, (2) collisions, injuries, or aggressive acts, (3) objects used unsafely, (4) visible damage or violent behaviour. Provide the bounding box coordinate of the most important region that can help determine whether this scene is Normal or Abnormal.}

  \textbf{Turn 2} (cropped region from Turn 1): \texttt{Based on the highlighted region, classify this surveillance scene as exactly one of: Abnormal or Normal. Respond with exactly one word first, then explain your reasoning.}
  \end{quote}

  The model generates a bounding box $[x_1, y_1, x_2, y_2]$ in Turn~1 (normalized to $[0,1000]$, rescaled to $[0,1]$), then classifies the cropped region in Turn~2. Any positive frame triggers an abnormal verdict. Focus boxes from Turn~1 are used for spatial grounding evaluation.

  \paragraph{SpatialVLM~\cite{chen2024spatialvlm}.}
  We implement a VQA Synth-inspired four-stage pipeline: (1)~YOLOv8x object detection, (2)~DepthPro monocular depth estimation per bounding box, (3)~structured spatial scene description fusing positions, boxes, depths, and pairwise relationships, (4)~Qwen3-VL-4B-Instruct classification. The Stage~4 prompt is:

  \begin{quote}\small\ttfamily
  You are a highly advanced Video Surveillance System. Analyze this frame using both visual evidence and the spatial scene analysis below. [\ldots] Using the spatial context, answer: Q1--Q5 covering unusual positions/depths, collision/aggression indicators, weapons, threatening proximity, and overall 3D layout. Respond with exactly ONE word first: Abnormal or Normal. Then explain your reasoning.
  \end{quote}

  We sample 6 frames for short videos and 10 for long ones. Inference uses bfloat16, greedy decoding (max 256 tokens); \texttt{<think>} blocks are stripped before parsing.

\paragraph{Evaluation Protocol for Grounded Baselines.}
For Kosmos-2, Visual-CoT, and SpatialVLM, we evaluate spatial grounding by extracting bounding boxes from model outputs and computing meanIoU and R@25 against GroundingDINO ground-truth boxes on the last frame of each test subclip. Classification predictions are extracted from the text output (``Normal''/``Abnormal'' verdict). All grounded baselines are evaluated zero-shot on UCF-Crime test data without any VAD-specific fine-tuning.

% =====================================================================
  \begin{figure*}[!htbp]
      \centering
      \includegraphics[width=\textwidth]{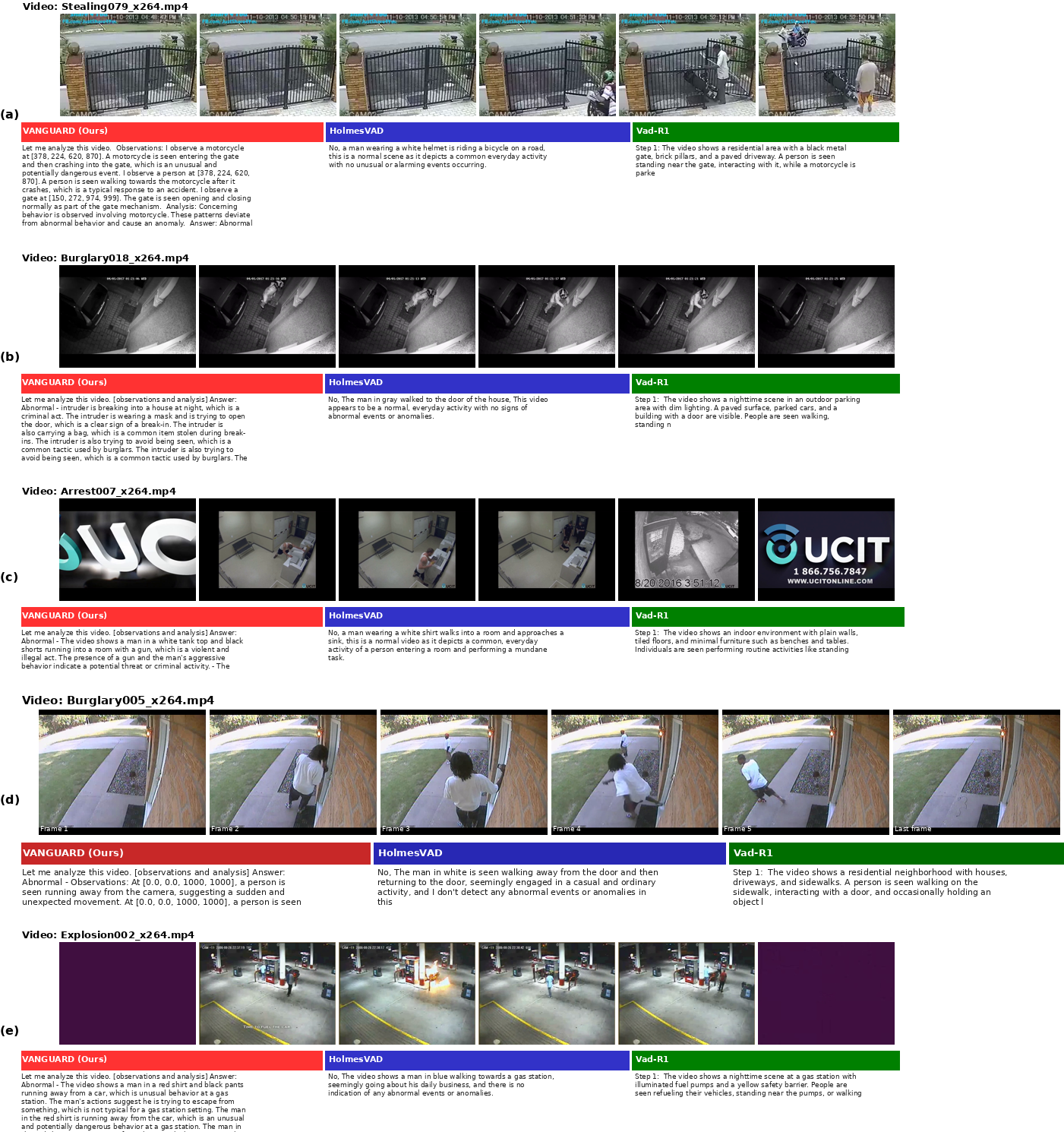}
      \caption{Qualitative comparison of CoT-grounded reasoning for video anomaly detection on UCF-Crime test samples. Each row shows sampled frames from a different
   test video. Columns display the generated reasoning from \textsc{Vanguard} (Ours), HolmesVAD, and Vad-R1. \textsc{Vanguard} produces structured, spatially-grounded reasoning---identifying anomalous objects with bounding box
  coordinates and per-object explanations (e.g., intruder breaking into a house at night in Burglary018 (b), person with a gun in Arrest007 (c))---while HolmesVAD
  generates brief responses that frequently misclassify anomalous scenes as normal, and Vad-R1 provides scene-level descriptions without explicit spatial grounding.
  The results demonstrate that \textsc{Vanguard}'s curriculum-trained CoT reasoning acts as both an interpretability mechanism and an implicit regularizer, producing
   more balanced and explainable predictions than classification-only or prompt-only baselines.}
      \label{fig:supp_reasoning_comparison}
  \end{figure*}

\section{Qualitative Comparisons}
\label{sec:supp_qualitative}

\subsection{Spatial Grounding}

Figure~\ref{fig:supp_spatial_comparison} provides side-by-side qualitative comparisons between grounded baselines and \textsc{Vanguard} on representative UCF-Crime test samples. Key observations:

\begin{itemize}
    \item \textbf{Kosmos-2} produces either no boxes or single coarse boxes covering the entire scene---its grounding tokens, trained on natural images, do not generalize to low-resolution, cluttered CCTV footage.
    \item \textbf{SpatialVLM} detects generic foreground objects regardless of anomaly relevance. For example, in Shoplifting044~(e) it localizes bystanders rather than the shoplifting act.
    \item \textbf{\textsc{Vanguard}} produces tight, object-specific boxes with correct anomaly attribution---localizing anomalous entities \emph{and} their context (e.g., fire and van in Arson009~(a), motorcycle on the ground in Stealing079~(d)). It is also more robust in low-visibility night scenes (Arson009~(a), Shooting026~(b)) where both baselines fail or yield imprecise localizations.
    \item Neither baseline connects spatial localization to anomaly semantics: they can find objects but cannot distinguish which ones are anomalous.
\end{itemize}

\subsection{Chain-of-Thought Reasoning}

  Figure~\ref{fig:supp_reasoning_comparison} presents qualitative comparisons of chain-of-thought reasoning between \textsc{Vanguard} and two reasoning-capable VAD
  baselines---HolmesVAD and Vad-R1---on representative UCF-Crime test samples. Each row displays sampled video frames alongside the generated reasoning text from all
  three methods. Key observations:
  \begin{itemize}
      \item \textbf{HolmesVAD} produces brief, surface-level responses (typically 1--2 sentences) that lack structured reasoning or spatial grounding. It frequently
  misclassifies abnormal videos as normal (e.g., Burglary018 (b), Explosion002 (e)), and when it does detect anomalies, it provides generic descriptions without
  localizing the anomalous objects or explaining the causal reasoning behind its judgment.
      \item \textbf{Vad-R1} employs a structured multi-step reasoning format (\texttt{<step1>}--\texttt{<step4>}) that separates scene description from anomaly
  recognition. However, its reasoning is limited to coarse scene-level descriptions and approximate spatial language (e.g., ``bottom left of the frame'') without
  producing explicit bounding box coordinates. Its outputs are also truncated in practice, limiting the depth of causal analysis.
      \item \textbf{\textsc{Vanguard}} generates detailed, spatially-grounded chain-of-thought reasoning that jointly localizes anomalous objects with normalized
  bounding box coordinates (e.g., \texttt{[x1, y1, x2, y2]} in $[0, 1000]$) and provides per-object anomaly attribution. Its structured format---\emph{Observations}
  $\rightarrow$ \emph{Analysis} $\rightarrow$ \emph{Answer}---connects spatial evidence to the final classification, yielding interpretable and verifiable
  predictions. For instance, in Stealing079 (a), \textsc{Vanguard} identifies a motorcycle entering through a gate with precise coordinates, while both baselines
  either miss the anomaly or provide vague descriptions.
  \end{itemize}

% =====================================================================
\section{VANGUARD-Bench: Additional Details}
\label{sec:supp_benchmark}

We detail the exact prompts used at each stage of the VANGUARD-Bench construction pipeline.
  
\begin{figure*}[t] \centering
\includegraphics[width=\textwidth]{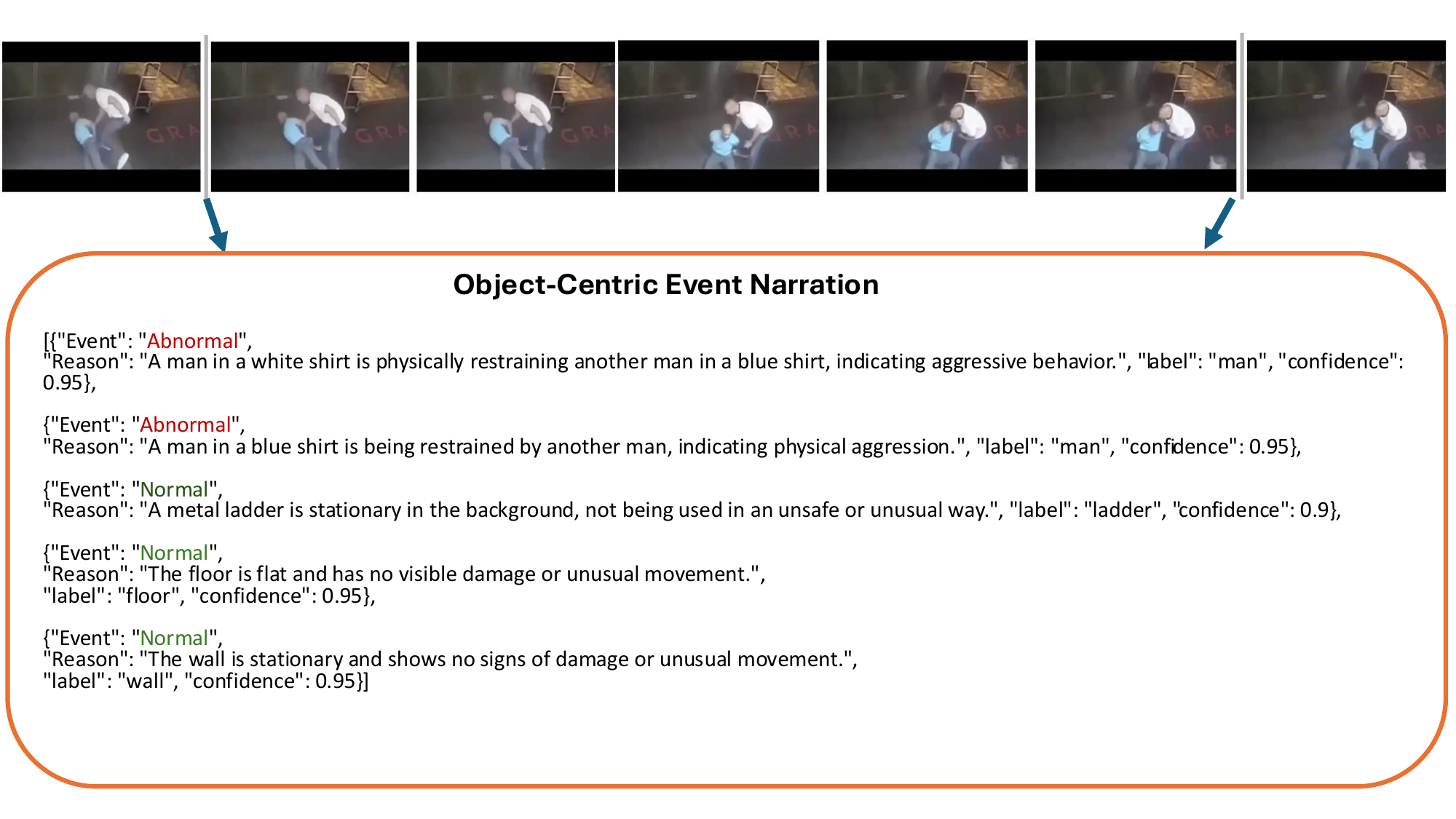} \caption{Pipeline output for a subclip from Arrest002. The VLM detects two men engaged in physical aggression (Abnormal) alongside passive scene objects---a ladder, floor, and wall (Normal). GroundingDINO localizes each object with bounding boxes.}
\label{fig:event_narration}
\end{figure*}

\begin{figure*}[t] \centering
\includegraphics[width=\textwidth]{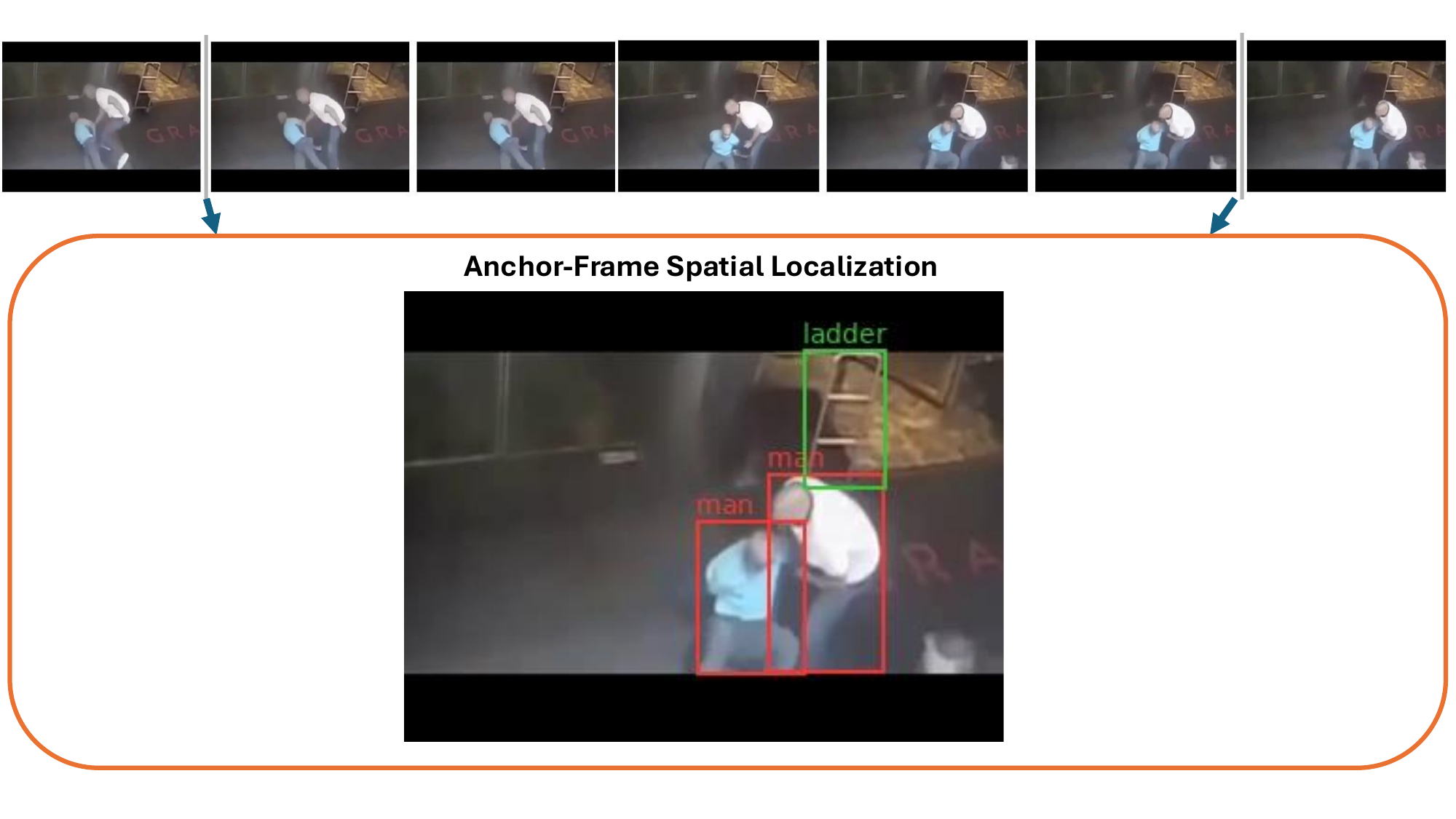} \caption{Spatial grounding output for an abnormal subclip from Arrest002. GroundingDINO localizes VLM-detected objects on the keyframe: two men involved in a physical altercation (red,
  Abnormal) and a stationary ladder in the background (green, Normal). Scene-level objects such as ``floor'' and ``wall'' were excluded as their bounding boxes exceeded 50\% of the frame area. A
  greedy deduplication step ensures each detected person is assigned a distinct bounding box.}
\label{fig:object_grounding}
\end{figure*}

\subsection{Object-Centric Event Narration.}
Given a subclip's frames and aligned temporal annotation sentences from the UCF-Crime dataset, we prompt Qwen3-VL-4B with the Object Detection Prompt:
  \begin{tcolorbox}[colback=gray!5, colframe=gray!50, title=Prompt for Object-Centric Event Narration, fonttitle=\bfseries\small, breakable]
  \small
  You are analyzing a surveillance video with temporal annotations. Ignore any biases emerging from text on the video. 
  \medskip
  
  \textbf{Temporal Annotations:} \texttt{\{annotations\_text\}}
  \medskip
  
  \textbf{Task 1: Binary Classification} --- Classify this video as Normal or Abnormal using the following questions:
  \begin{enumerate}[nosep, leftmargin=1.5em]
      \item Are there any people not in their typical positions or engaging in activities inconsistent with usual behavior?
      \item Are there any collisions between people, vehicles, or objects that indicate abnormal or unsafe behavior?
      \item Are there any injuries visible (e.g., person lying on the ground, limping, requiring assistance)?
      \item Is there any abuse or aggressive behavior (e.g., pushing, hitting, kicking)?
      \item Are there any objects or equipment being used in an unsafe or unusual way?
      \item Is there any visible damage or unusual movement that indicates an anomaly?
      \item Are there any signs of physical aggression, fighting, or violent behavior between people?
  \end{enumerate}
  \medskip
  \textbf{Task 2: Object Detection} --- List ALL detected objects in the video. For each object return:
  \begin{itemize}[nosep, leftmargin=1.5em]
      \item The object name (in lowercase)
      \item A confidence score between 0 and 1
      \item Event: ``Normal'' if the object is behaving normally, ``Abnormal'' ONLY if the object is DIRECTLY involved in abnormal/dangerous activity
      \item Reason: a short factual description of what the object is doing. The Reason MUST be consistent with the Event --- if the Reason describes normal/routine
  behavior, the Event MUST be ``Normal''. Only set Event to ``Abnormal'' when the Reason clearly describes harmful, violent, or dangerous behavior.
  \end{itemize}
  \medskip
  \textbf{CRITICAL:} You MUST respond ONLY with a valid JSON array. Do not include any text before or after the JSON.
  \medskip
  Output the results as a JSON array, where each element is:
  \begin{lstlisting}[basicstyle=\small\ttfamily, backgroundcolor=\color{gray!10}, breaklines=true, breakatwhitespace=false]
  {"Event": "Normal" or "Abnormal",
   "Reason": "short factual description",
   "label": "detected object in lowercase",
   "confidence": float}
  \end{lstlisting}
  \end{tcolorbox}

The VLM processes the full subclip video as visual input alongside a set of binary classification questions and object detection instructions, leveraging both spatial appearance and temporal dynamics to produce per-object event narrations. Each detected object receives an independent Normal/Abnormal event tag with a short factual reason grounded in visual evidence. Crucially, the event label must be consistent with the stated reason---objects exhibiting routine behavior are tagged Normal even in abnormal videos, ensuring that only objects \textit{directly involved} in anomalous activity receive the Abnormal designation. An example output is shown in Fig.~\ref{fig:event_narration}.

\subsection{Anchor-Frame Spatial Localization.}
For each detected object from the event narration stage, we query GroundingDINO~\cite{liu2024grounding} with \textit{two} text prompts: (i)~the object label (e.g., ``man'') and (ii)~the causal reason produced by the VLM (e.g., ``physically restraining another man''). Providing the reason as an additional query improves recall, as GroundingDINO can leverage the richer semantic description to detect objects that a label alone might miss. Both prompts are passed jointly with a box confidence threshold of $0.25$. Rather than grounding on a single frame, we evaluate all objects jointly on each candidate frame, starting from the last frame of the subclip and falling back to up to five earlier frames sampled uniformly in reverse. For each frame, GroundingDINO returns detections for all objects, and a greedy deduplication step assigns one unique bounding box per object: detections are ranked by confidence and assigned in order, with any candidate whose IoU with an already-assigned box exceeds $0.5$ skipped in favor of the next-best detection. This prevents multiple VLM objects sharing the same label (e.g., two instances of ``man'') from collapsing onto a single detection. The frame that successfully grounds the most objects is selected, ensuring all objects share a consistent spatial reference. We retain only detections whose
  label matches the original object label via bidirectional substring matching; detections triggered by the reason string but not matching the object label are discarded, ensuring the reason serves purely as a contextual grounding aid. After assignment, any bounding box covering more than $50\%$ of the frame area is removed, as scene-level detections (e.g., ``floor'', ``wall'') do not provide meaningful localization signal. An example is shown in Fig.~\ref{fig:object_grounding}. In total, this achieves a $92.5\%$ grounding rate ($147{,}067$ of $159{,}008$ objects).

\begin{table}[t]
  \centering
  \caption{Phase~1: Subclipping strategy statistics. Videos are segmented via CLIP-based keyframe extraction and subsampled to at most two subclips per video.}
  \label{tab:phase1-subclipping}
  \begin{tabular}{lr}
  \toprule
  \textbf{Metric} & \textbf{Value} \\
  \midrule
  Videos (Normal / Abnormal)              & 1{,}288 (644 / 644) \\
  Total subclips                          & 2{,}092 \\
  \quad Normal / Abnormal                 & 1{,}012 / 1{,}080 \\
  Subclips / video (mean / median)        & 1.9 / 2 \\
  Subclip duration, median                & 21.7\,s \\
  \quad p25 / p75                         & 5.5\,s / 69.5\,s \\
  \quad $\leq$ 10\,s                      & 35.2\% \\
  \quad $\leq$ 30\,s                      & 57.9\% \\
  \quad $\leq$ 60\,s                      & 71.8\% \\
  \bottomrule
  \end{tabular}
  \end{table}

  \begin{table}[t]
  \centering
  \caption{Phase~2: Object grounding statistics. Up to ten image-grounded subclips are sampled per video, each annotated with VLM-based object detection and GroundingDINO spatial localization on a single anchor frame.}
  \label{tab:phase2-grounding}
  \resizebox{\columnwidth}{!}{%
  \begin{tabular}{lr}
  \toprule
  \textbf{Metric} & \textbf{Value} \\
  \midrule
  Total samples (Normal / Abnormal)       & 7{,}807 (3{,}941 / 3{,}866) \\
  Subclips / video (mean / median)        & 6.1 / 7 \\
  Total object instances                  & 29{,}203 \\
  Unique object labels                    & 2{,}606 \\
  \quad Singleton labels                  & 1{,}412 (54.2\%) \\
  \quad Rare labels (2--5 occurrences)    & 729 (28.0\%) \\
  Objects / subclip (mean / median)       & 3.7 / 3 \\
  Grounded with box (mean / median)       & 2.6 / 3 \\
  Grounding success rate                  & 74.6\% \\
  Subclips with no grounding              & 19.7\% \\
  \midrule
  \multicolumn{2}{l}{\textit{Most frequent objects}} \\
  \quad person / car / man                & 2{,}996 / 1{,}792 / 1{,}518 \\
  \quad woman / motorcycle / counter      & 638 / 557 / 549 \\
  \quad chair / camera / timestamp        & 547 / 501 / 408 \\
  \midrule
  \multicolumn{2}{l}{\textit{Highest ungrounded rate ($\geq$20 instances)}} \\
  \quad picture                           & 86\% (25/29) \\
  \quad ground                            & 83\% (19/23) \\
  \quad lighting                          & 73\% (90/123) \\
  \quad fan / driveway / house            & 72\% / 71\% / 70\% \\
  \quad surveillance camera               & 65\% (44/68) \\
  \quad trash can / parking lot           & 64\% / 64\% \\
  % \midrule
  % \multicolumn{2}{l}{\textit{Most frequent ungrounded (absolute count)}} \\
  % \quad car / person / camera             & 635 / 536 / 290 \\
  % \quad man / counter / chair             & 276 / 271 / 250 \\
  % \quad motorcycle / door / timestamp     & 212 / 196 / 192 \\
  \bottomrule
  \end{tabular}%
  }% end resizebox
  \end{table}

  \begin{table}[t]
  \centering
  \caption{Phase~3: Chain-of-thought annotation statistics. CoT text is generated
  by converting VLM object detections and bounding
  boxes into a structured Observations--Analysis--Answer format.}
  \label{tab:phase3-cot}
  \resizebox{\columnwidth}{!}{%
  \begin{tabular}{lr}
  \toprule
  \textbf{Metric} & \textbf{Value} \\
  \midrule
  Total CoT annotations                   & 2{,}092 \\
  \quad Answer: Normal / Abnormal         & 1{,}012 / 1{,}080 \\
  Section completeness (Obs / Ana / Ans)  & 98.2 / 98.2 / 100\% \\
  \midrule
  \multicolumn{2}{l}{\textit{Overall length}} \\
  \quad Words (mean / median)             & 149 / 132 \\
  \quad Characters (mean / median)        & 915 / 818 \\
  \quad p25 / p75 (words)                 & 99 / 170 \\
  \midrule
  \multicolumn{2}{l}{\textit{Section breakdown}} \\
  \quad Observations, words (mean / median) & 117 / 108 \\
  \quad Analysis, words (mean / median)   & 15 / 11 \\
  \midrule
  \multicolumn{2}{l}{\textit{Spatial content}} \\
  \quad Objects described / CoT (mean / median) & 4.6 / 4 \\
  \quad Bbox coordinates / CoT (mean / median)  & 4.0 / 4 \\
  \quad CoTs with $\geq$1 bbox reference        & 98.2\% \\
  \bottomrule
  \end{tabular}%
  }% end resizebox
  \end{table}

\subsection{Chain-of-Thought Synthesis.}

% converts raw VLM object annotations into structured training text. It takes the JSON annotation (list of detected objects with Event/Reason/label/bounding boxes), the ground-truth label, and optional frame dimensions, then produces a three-section CoT string: 1. Parsing: Accepts annotation as a JSON string or list. Strips code fences and <think> blocks. If already a CoT response (contains "Answer:"), returns as-is. 2. Observations: For each object, builds a phrase: a {label} at [{x1}, {y1}, {x2}, {y2}] ({reason}). Bounding boxes are normalized to [0, 1000] if frame_hw is provided, otherwise written as pixel coordinates. All objects are joined with semicolons. 3. Analysis: Templated based on the ground-truth label. If Abnormal with identified abnormal objects, names the concerning objects specifically.  If Abnormal with no flagged objects, gives a generic abnormality statement. If Normal, states routine behavior. 4. Output: Returns the full CoT string: Let me analyze this video. Observations: I observe a person at [135, 122, 164, 155] (walking normally); a knife at [400, 300, 450, 380] (being held aggressively). Analysis: Concerning behavior is observed involving knife. These patterns deviate from typical surveillance activity and indicate an anomaly. Answer: Abnormal

To produce the final instruction-tuning data, each subclip's per-object annotations from the event narration and spatial localization stages are converted into a structured object context. Each object is
  represented as a bullet point containing its label, bounding box coordinates (normalized to $[0, 1000]$), causal reason, and event tag---for example: \texttt{- man at [247, 318, 448, 853]: "aggressive posture" (Abnormal)}. This structured object context, together with the subclip's temporal annotation sentences and the ground-truth video label, is passed to Qwen3-VL-4B with the Chain-of-Thought Generation Prompt:
\begin{tcolorbox}[colback=blue!3, colframe=blue!40, title=Chain-of-Thought Generation Prompt, fonttitle=\bfseries\small, breakable]
  \small
  You are analyzing a surveillance video. The following objects were detected:
  
  \texttt{\{object\_context\}}
  \medskip
  
  \textbf{Temporal Annotations:} \texttt{\{annotations\_text\}}

  \medskip
  Coordinates are normalized to [0, 1000] where (0,0) is top-left and (1000,1000) is bottom-right.

  The video is labeled: \texttt{\{label\}}

  \medskip
  Write a chain-of-thought analysis. Do NOT start with ``Let me analyze this video'' --- that prefix will be added automatically.

  \medskip
  \textbf{Structure:}
  \begin{itemize}[nosep, leftmargin=1.5em]
      \item \textbf{Observations:} Describe 2--3 key behaviors you observe, referencing object locations with bounding box coordinates $[x_1, y_1, x_2, y_2]$ in $[0, 1000]$ range where
  available.
      \item \textbf{Analysis:} Explain your reasoning for classifying this video as \texttt{\{label\}} with specific details from the observations.
      \item End with exactly: \texttt{Answer: \{label\}}
  \end{itemize}

  \medskip
  \textbf{IMPORTANT:} Write as if you are directly watching the video. Use phrases like ``In the video, I observe\ldots'' --- do NOT reference temporal annotations, context, input, or any
  information source. Everything must sound like first-person visual observation.

  \medskip
  Be specific and grounded in the detected objects. 3--5 sentences total.
  \end{tcolorbox}
  
The model is instructed to write a first-person visual analysis using exactly three sections: \textbf{Observations}, referencing object locations with bounding box coordinates; \textbf{Analysis}, explaining the reasoning for the classification; and \textbf{Answer}, stating the final Normal/Abnormal label. Importantly, the model is required to write as if directly observing the video---it must not reference the temporal annotations or any input context, ensuring the generated CoT reads as genuine visual reasoning.

An example of the generated CoT for the subclip shown in Fig.~\ref{fig:event_narration} and Fig.~\ref{fig:object_grounding} is given below:

\begin{tcolorbox}[colback=green!3, colframe=green!40, title=Generated Chain-of-Thought Response, fonttitle=\bfseries\small, breakable]
  \small
  Let me analyze this video.

  \medskip
  \textbf{Observations:} I observe a man in a white shirt physically restraining another man in a blue shirt, with the blue-shirted man's bounding box at $[456, 559, 634, 849]$, indicating forceful contact. The
   white-shirted man's action is clearly aggressive, as he holds the blue-shirted man in place. A metal ladder at $[661, 131, 804, 455]$ remains stationary and uninvolved, while the floor and wall show no signs
   of disturbance.

  \medskip
  \textbf{Analysis:} The physical restraint between the two men is a clear indicator of violent behavior, which deviates from normal activity in the scene. The presence of aggressive physical interaction,
  combined with the stationary background elements, reinforces that the event is anomalous and potentially dangerous. No other objects or movements suggest a non-violent context.

  \medskip
  \textbf{Answer: Abnormal}
  \end{tcolorbox}

\subsection{Dataset Statistics}
We randomly sample 80\% of the UCF-Crime training split to form \textsc{Vanguard-Bench}, yielding 1{,}288 videos (644 normal, 644 abnormal). CLIP-based keyframe extraction produces 29{,}041 subclips (median 21.7\,s), subsampled to two per video for a balanced set of 2{,}092 training subclips
  (Table~\ref{tab:phase1-subclipping}). For spatial supervision, up to ten anchor-frame subclips per video are annotated with VLM-based object detection and
  GroundingDINO localization, yielding 7{,}807 image-grounded samples with 29{,}203 object instances across 2{,}606 unique labels at a 74.6\% grounding rate (Table~\ref{tab:phase2-grounding}). Each subclip is paired with a structured
  chain-of-thought annotation in Observations--Analysis--Answer format, averaging 149 words per CoT (Table~\ref{tab:phase3-cot}).

% =====================================================================
\section{Training Details}
\label{sec:supp_implementation}

\subsection{Model Architecture Details}

\begin{itemize}
    \item \textbf{Base model}: Qwen3-VL-4B-Instruct (4B parameters).
    \item \textbf{LoRA}: Rank $r{=}64$, $\alpha{=}16$, dropout $0.1$. In Stages~1 and~3, LoRA is applied to the language model's query and value projections (\texttt{q\_proj}, \texttt{v\_proj}). In Stage~2, LoRA is additionally applied to the vision encoder's fused QKV and output projections (\texttt{attn.qkv}, \texttt{attn.proj}) across all 24 ViT blocks, enabling the vision encoder to co-adapt its spatial features for bounding box prediction during the spatial grounding stage.
    \item \textbf{Projection head}: Two-layer MLP ($2560 \to 2560$, ReLU, $2560 \to 128$) mapping pooled hidden states to a 128-dimensional feature space.
    \item \textbf{Classifier head}: Dropout ($p{=}0.5$) followed by linear ($128 \to 1$) producing a binary anomaly logit.
    \item \textbf{Pooling}: Attention-mask-weighted mean pooling over the last layer's hidden states.
    \item \textbf{Generation}: Maximum 512 new tokens, using the base model's \texttt{generate()} method.
\end{itemize}

\subsection{Training Data Format and Curriculum Prompts}
\paragraph{Model Input.}
  Each training sample consists of a multi-turn chat-format conversation:
  \begin{itemize}
  \item \textbf{System message}: Task-specific instructions (varies by curriculum stage; see below).
     \item \textbf{User message}: Depends on the stage:
      \begin{itemize}
          \item \textit{Stages 1 \& 3}: Video/clip frames embedded as visual tokens.
          \item \textit{Stage 2}: A single image (last frame of the subclip) followed by:
          \begin{quote}\small\ttfamily
          Locate every instance that belongs to the following categories: \{labels\}. Output bounding box coordinates in JSON for every label.
          \end{quote}
      \end{itemize}

      \item \textbf{Assistant response} (training target):
      \begin{itemize}
          \item \textit{Stage 1}: None---the classifier head is trained directly on the binary label; no text generation loss is applied.
          \item \textit{Stage 2}: A JSON array of detections:
          \begin{quote}\small\ttfamily
          [\{"bbox\_2d": [x1, y1, x2, y2], "label": "CATEGORY"\}, ...]
          \end{quote}
          with coordinates normalized to $[0, 1000]$.
          \item \textit{Stage 3}: Structured chain-of-thought text with three sections---\\
          \texttt{Observations} (per-object descriptions with bounding box coordinates), \texttt{Analysis} (reasoning summary), and
   \texttt{Answer: Normal/Abnormal}.
      \end{itemize}
  \end{itemize}

The system messages for each curriculum stage are as follows:

\textit{Stage 1 (Classifier Warmup):}
\begin{quote}\small\ttfamily
  You are a surveillance video analysis expert. Classify the video as Normal or Abnormal strictly using visual evidence. You MUST end your response with `Answer: Abnormal' or `Answer:
  Normal'. \end{quote}
  
\textit{Stage 2 (Vanguard-Bench Spatial Grounding):}
\begin{quote}\small\ttfamily
  You are a precise object detector for surveillance footage. Given an image (the last frame of a video clip), locate every instance of the specified object categories and predict their bounding boxes.
  Output ONLY valid JSON --- no extra text, no markdown, no code fences.
  Format: [\{"bbox\_2d": [x1, y1, x2, y2], "label": "CATEGORY", "anomaly": true/false, "reason": "brief explanation of why this object is normal or anomalous"\}, ...]
  Coordinates are integers in [0,\,1000] where (0,\,0) is top-left and (1000,\,1000) is bottom-right. Each bounding box must tightly fit the object --- boxes should NOT cover the
  entire frame and must not exceed 60\% of the frame area. If a category is not visible, omit it. If nothing is visible, output [].
  \end{quote}

\textit{Stages~2 \& 3 (Vanguard-Bench CoT Reasoning):} Both stages share the following prompt:
\begin{quote}\small\ttfamily
  You are analyzing a surveillance video.
  Coordinates are normalized to [0,\,1000] where (0,\,0) is top-left and (1000,\,1000) is bottom-right. Write a chain-of-thought analysis. You MUST use EXACTLY this format with these exact section headers:
  Observations: Describe 2--3 key behaviors you observe, referencing object locations with bounding box coordinates [x1, y1, x2, y2] in [0,\,1000] range where available.
  Analysis: Explain your reasoning for classifying this video as Normal or Abnormal with specific details from the observations.
  Answer: Normal or Abnormal
  IMPORTANT: Be specific and grounded in the detected
  objects. 3--5 sentences total across Observations and Analysis.
  \end{quote}

\begin{table}[t]
\centering
\caption{Training hyperparameters per curriculum stage.}
\label{tab:supp_training_config}
\resizebox{\columnwidth}{!}{%
\begin{tabular}{lccc}
\toprule
\textbf{Hyperparameter} & \textbf{Stage 1} & \textbf{Stage 2} & \textbf{Stage 3} \\
\midrule
Epochs & 2 & 3 & 3 \\
Peak learning rate & $1 \times 10^{-3}$ & $5 \times 10^{-4}$ & $5 \times 10^{-4}$ \\
LR scheduler & Cosine & Cosine & Cosine \\
Warmup ratio & 0.1 & 0.05 & 0.05 \\
Batch size (per device) & 4 & 4 & 4 \\
Gradient accumulation steps & 2 & 2 & 2 \\
Effective batch size & 8 & 8 & 8 \\
Max gradient norm & 100 & 100 & 100 \\
Weight decay & 0.01 & 0.01 & 0.01 \\
Optimizer & AdamW & AdamW & AdamW \\
$\beta_1, \beta_2$ & 0.9, 0.999 & 0.9, 0.999 & 0.9, 0.999 \\
$\epsilon$ & $10^{-6}$ & $10^{-6}$ & $10^{-6}$ \\
Precision & bfloat16 & bfloat16 & bfloat16 \\
\midrule
Trainable parameters & Heads only & LoRA + Heads & LoRA + Heads \\
\bottomrule
\end{tabular}%
}% end resizebox
\end{table}

\subsection{Training Hyperparameters}

Table~\ref{tab:supp_training_config} details the training configuration for each curriculum stage.

\paragraph{Cross-stage LR continuity.}
To avoid abrupt learning rate discontinuities when transitioning between stages, each stage's warmup begins from the cosine-decayed endpoint of the previous stage (approximately $\text{peak\_lr} \times 0.01$). This ensures smooth optimization dynamics across stage boundaries.

\subsection{Loss Weight Configurations ($\lambda$ Values)}
\label{sec:supp_lambda}

Table~\ref{tab:supp_lambda} specifies the loss weights used at each curriculum stage. The key design principle is progressive introduction: Stage~1 uses only classification loss, Stage~2 adds text generation and spatial GIoU, and Stage~3 drops GIoU while retaining text generation for CoT refinement.

\paragraph{Rationale for $\lambda$ choices.}
\begin{itemize}
    \item $\lambda_\text{bce} = 1.0$ throughout: The classification signal anchors training at every stage. Keeping it constant prevents the classifier from drifting during text generation fine-tuning.
    \item $\lambda_\text{lm} = 0.0 \to 0.5$: Text generation loss is introduced at 0.5 in Stages~2--3. We found that $\lambda_\text{lm} < 0.5$ in Stage~2 led to poor recall of objects in the generated descriptions, while $\lambda_\text{lm} > 0.5$ degrades classification performance as the model prioritizes text fluency over discriminative features.
    \item $\lambda_\text{giou} = 1.0$ in Stage~2 only: $\mathcal{L}_\text{lm}$ is effective for learning how many objects to detect but treats each coordinate digit independently, making it insufficient for accurate box regression. The GIoU loss adds geometry-aware supervision. As shown in Table~\ref{tab:supp_lambda_sensitivity}, $\lambda_\text{giou} = 1.0$ yields the best meanIoU and AUC; both lower and higher values degrade performance. Stage~3 drops GIoU as spatial grounding has been internalized.
\end{itemize}

\begin{table}[t]
\centering
\caption{Loss weight ($\lambda$) configuration per curriculum stage.}
\label{tab:supp_lambda}
\small
\begin{tabular}{lccc}
\toprule
\textbf{Loss weight} & \textbf{Stage 1} & \textbf{Stage 2} & \textbf{Stage 3} \\
\midrule
$\lambda_\text{bce}$ (Binary CE) & 1.0 & 1.0 & 1.0 \\
$\lambda_\text{lm}$ (Text generation) & 0.0 & 0.5 & 0.5 \\
$\lambda_\text{giou}$ (Text-coord GIoU) & 0.0 & 1.0 & 0.0 \\
\midrule
Data: Image-level detection & -- & 80\% & -- \\
Data: UCF-Train (Videos-label) & 100\% & -- & -- \\
Data: Vanguard-Bench & -- & 20\% & 100\% \\
\bottomrule
\end{tabular}
\end{table}

\begin{table}[t]
\centering
\caption{Sensitivity analysis of GIoU loss weight on UCF-Crime validation set ($\lambda_\text{bce}=1.0$, $\lambda_\text{lm}=0.5$ fixed). Bold indicates the configuration used in our final model. Note that $^\ast$AUC and F1 are computed at the subclip level in Vanguard-Bench.}
\label{tab:supp_lambda_sensitivity}
\small
\begin{tabular}{cccccc}
\toprule
 $\lambda_\text{lm}$ & $\lambda_\text{giou}$ & AUC$^\ast$ & F1$^\ast$ & meanIoU & Recall@25 \\
\midrule
 0.1 & 0.5 & 0.9553 & 0.9038 & 0.5705 & 0.4732 \\
 0.5 & 0.5 & 0.9553 & 0.9038 & 0.5705 & 0.4732 \\
\textbf{0.5} & \textbf{1.0} & \textbf{0.9635} & \textbf{0.9016} & \textbf{0.6191} & \textbf{0.5061} \\
 0.5 & 2.0 & 0.9551 & 0.8927 & 0.5805 & 0.4792 \\
 0.5 & 3.0 & 0.9523 & 0.8875 & 0.5764 & 0.4703 \\
\bottomrule
\end{tabular}
\end{table}

% =====================================================================
\section{Ablation Studies}
\label{sec:supp_ablations}

\subsection{Curriculum Training vs.\ Joint Optimization}
\label{sec:supp_curriculum_ablation}
To isolate the effect of curriculum staging, we train a single-phase baseline that applies all losses simultaneously. Starting from the Stage~1 classifier checkpoint, the model is trained for 3 epochs at a learning rate of $5 \times
  10^{-4}$ on a 50/50 mixture of image-level detection samples (7{,}807 samples) and video-level CoT samples (2{,}092 samples). All losses---binary cross-entropy ($w{=}1.0$), text
  generation ($w{=}0.5$), and GIoU ($w{=}1.0$)---are active from the first
  step, with LoRA adapters and classifier unfrozen throughout. Each sample
  receives a modality-appropriate system prompt: image inputs use the Stage~2
  detection prompt requesting JSON-format bounding box output, while video
  inputs use the Stage~3 CoT prompt eliciting structured
  Observations--Analysis--Answer reasoning. This setup tests whether the
  curriculum's staged introduction of losses and data modalities provides a
  benefit over joint training on the same total data.

\begin{table*}[t]
\centering
\caption{Curriculum training vs.\ joint optimization on UCF-Crime. The curriculum achieves substantially stronger spatial grounding than monolithic training, with competitive classification performance.}
\label{tab:supp_curriculum}
\small
\begin{tabular}{lcccccc}
\toprule
\textbf{Training Strategy} & \textbf{AUC} & \textbf{F1} & \textbf{Prec.} & \textbf{Rec.} & \textbf{meanIoU} & \textbf{R@25} \\
\midrule
Stage 1 only (2 ep.) & \underline{0.9436} & 0.7750 & \underline{0.8235} & 0.7317 & -- & -- \\
Stage 1 $\to$ Joint (all losses, 2 + 6 ep.) & \textbf{0.9670} & \textbf{0.8782} & \textbf{0.9084} & 0.8500 & 0.2064 & 0.089 \\
Stage 1 $\to$ 2 (2+3 ep.) & 0.9298 & 0.8104 & 0.7892 & 0.8333 & 0.58 & 0.46 \\
Stage 1 $\to$ 2 $\to$ 3 (2+3+3 ep.) & 0.9378 & \underline{0.8360} & 0.8108 & \textbf{0.8627} & \textbf{0.62} & \textbf{0.51} \\
\bottomrule
\end{tabular}
\end{table*}

\paragraph{Analysis.} Joint training achieves the highest raw classification metrics (AUC 0.9670, F1 0.8782) but catastrophically poor spatial grounding (meanIoU 0.2064, R@25 0.089)---training all losses simultaneously allows classification and text generation gradients to overwhelm the spatial objective. The three-stage curriculum recovers grounding ($3\times$ higher meanIoU, $5.7\times$ higher R@25) with comparable classification (AUC 0.9378, F1 0.8360). Stage~3 refines both axes over Stage~2: AUC improves by 0.8pp, F1 by 2.6pp, meanIoU by 4pp, and R@25 by 5pp, confirming that the CoT-only stage consolidates classification and spatial skills learned in earlier stages.

\subsection{Effect of Number of Training Subclips}
\label{sec:supp_subclip_ablation}

Each UCF-Crime video is decomposed into multiple subclips via CLIP-based scene segmentation. We ablate the number of retained subclips per video to understand how training data diversity affects model performance.

Increasing from 2 to 4 subclips per video, improvements are marginal. This saturation occurs because additional subclips from the same video share similar visual characteristics and provide diminishing novel information. For computational efficiency, we use 2 subclips per video in our default configuration, which offers the best trade-off between training cost and performance.

\begin{table}[t]
\centering
\caption{Effect of subclips per video on UCF-Crime performance (Stage~3 trained with different subclips/video). Increasing subclips improves performance up to a saturation point.}
\label{tab:supp_subclips}
\small
\begin{tabular}{cccc}
\toprule
\textbf{Subclips/video} & \textbf{Train samples} & \textbf{AUC} & \textbf{F1} \\
\midrule
%1 & $\sim$1,610 & 0.9012 & 0.7241 & 0.48 & 0.35 \\
2 & $2{,}092$ & {0.9378} & {0.8360} \\
4 & $\sim$4{,}200 & \textbf{0.9429} & \underline{0.8509} \\
%6 & $\sim$6{,}300 & 0.9065 & 0.7550 \\
%4 & $\sim$4,200 & 0.9341 & 0.8218 & 0.60 & 0.49 \\
\bottomrule
\end{tabular}
\end{table}

% =====================================================================
\section{Additional Experimental Details}
\label{sec:supp_additional}

\subsection{Evaluation Metrics: Spatial Grounding}
We evaluate the quality of predicted bounding boxes against ground-truth annotations derived from GroundingDINO~\cite{liu2024grounding} on the last frame of each subclip. For each predicted box, we compute the best-match Intersection-over-Union (IoU) against all ground-truth boxes using greedy assignment, and report the mean across all matched pairs (\textbf{meanIoU}). Detection recall is measured at an IoU threshold of 0.25 (\textbf{R@25}), representing the fraction of ground-truth boxes matched above this threshold. These metrics are computed only on abnormal samples with annotated spatial grounding; normal samples contribute no boxes by design.

\subsection{Seed and Reproducibility}
All experiments use seed 42 for both data shuffling and model initialization. We use deterministic cuDNN operations where available. Results may vary slightly across GPU architectures due to non-deterministic atomics in \texttt{bfloat16} reductions.

\section{Real-time Video Processing}
\label{sec:supp_inference}
For real-time deployment, \textsc{Vanguard} applies the same CLIP-based scene segmentation used during dataset construction: every 15th frame is encoded with CLIP ViT-B/32, and a subclip boundary is declared when the cosine similarity to the previous boundary drops below $\tau{=}0.92$. The accumulated segment is then passed to the model, which returns a binary anomaly prediction, a chain-of-thought explanation with bounding box coordinates, and a confidence score. This event-driven design invokes the model only on scene changes rather than at fixed intervals, naturally adapting inference frequency to scene dynamics.

The threshold $\tau$ serves as a deployment-time knob: lower values (e.g., $0.85$) yield fewer, longer subclips and reduce compute but may group distinct events; higher values (e.g., $0.95$) catch subtler changes at greater cost. We find $\tau{=}0.92$ to be a robust default. The CLIP gating overhead is negligible (${\sim}$5\,ms/frame) relative to the VLM forward pass (${\sim}$1--2\,s/subclip), so total cost scales with the number of triggered subclips rather than frame rate.

\begin{table}[t]
\centering
\caption{Per-category anomaly detection accuracy on the UCF-Crime test set. Categories involving abrupt visual events achieve near-perfect recall, while Shoplifting---a visually subtle, slow-onset category---drops to 66.7\%. Normal video accuracy (72.7\%) reflects a 27.3\% false-positive rate.}
\label{tab:supp_per_category}
\small
\begin{tabular}{lrrrrr}
\toprule
\textbf{Category} & \textbf{Total} & \textbf{TP} & \textbf{FN} & \textbf{Acc.} & \textbf{Recall} \\
\midrule
Abuse             &   2 &   2 &  0 & 1.000 & 1.000 \\
Arrest            &   5 &   5 &  0 & 1.000 & 1.000 \\
Arson             &   9 &   9 &  0 & 1.000 & 1.000 \\
Assault           &   3 &   3 &  0 & 1.000 & 1.000 \\
Burglary          &  13 &  12 &  1 & 0.923 & 0.923 \\
Explosion         &  21 &  20 &  1 & 0.952 & 0.952 \\
Fighting          &   5 &   5 &  0 & 1.000 & 1.000 \\
RoadAccidents     &  23 &  23 &  0 & 1.000 & 1.000 \\
Robbery           &   5 &   5 &  0 & 1.000 & 1.000 \\
Shooting          &  23 &  22 &  1 & 0.957 & 0.957 \\
Shoplifting       &  21 &  14 &  7 & 0.667 & 0.667 \\
Stealing          &   5 &   5 &  0 & 1.000 & 1.000 \\
Vandalism         &   5 &   5 &  0 & 1.000 & 1.000 \\
\midrule
Normal            & 150 & \multicolumn{2}{c}{109 TN / 41 FP} & 0.727 & -- \\
\bottomrule
\end{tabular}
\end{table}

\section{Biases from VANGUARD-Bench}
\label{sec:supp_biases}

Several design choices in the VANGUARD-Bench construction pipeline introduce systematic biases that propagate into the trained model.

\paragraph{Subclipping strategy bias.}
CLIP-based scene segmentation ($\tau{=}0.92$) reliably isolates anomalies that produce abrupt visual discontinuities (explosions, fights, collisions) but struggles with slow-onset events such as shoplifting or gradual arson, which unfold within a visually homogeneous segment and get merged with surrounding normal footage. The model consequently receives cleaner supervision for visually dramatic categories and noisier signals for subtle ones. As shown in Table~\ref{tab:supp_per_category}, categories involving sudden visual change---Arson, Explosion, Fighting, RoadAccidents---achieve perfect or near-perfect recall, while Shoplifting, whose anomalous behavior unfolds gradually within a visually static retail scene, drops to 66.7\% recall (7 of 21 missed). This gap suggests that CLIP-based boundaries fail to isolate the relevant shoplifting segments during training data construction, resulting in noisier supervision for this category.

% \paragraph{Subclip cap and training distribution.} Retaining only two of a median seven subclips per video biases training toward the most visually salient anomaly instance, discarding secondary or subtler segments. Videos with multiple anomalous episodes (e.g., separate confrontation and escape phases) are only partially represented. As shown in Table~\ref{tab:supp_subclips}, increasing the subclip cap improves both classification and grounding up to a saturation point, confirming this as a diversity--cost trade-off.

\paragraph{Last-frame anchor bias.}
Spatial grounding annotations are derived from GroundingDINO on the \emph{last frame} of each subclip. This is a reasonable default---objects are grounded while still visible, before they exit the scene---and works well for culminating events (e.g., a completed assault, a fully developed fire). However, single-frame anchoring biases the model toward end-of-segment object positions. Multi-frame anchoring, where boxes are extracted at several keyframes within a subclip and aggregated, could provide richer spatial supervision and improve generalization to mid-event localization.

\paragraph{Teacher model cascading errors.}
The annotation pipeline relies on GroundingDINO for spatial localization, which achieves 92.5\% grounding overall but drops sharply for rare or small categories (``weapon'' AP@50 0.000, ``fire'' AP@50 0.258), imposing a category-dependent accuracy ceiling on the trained model. Qwen3-VL's contribution is more constrained---it selects and narrates from manually provided UCF-Crime annotations rather than generating labels from scratch---so its bias impact is limited to how it synthesizes the given context into CoT responses.

\paragraph{Geographic and demographic bias.}
Finally, VANGUARD-Bench inherits UCF-Crime's geographic skew (predominantly US surveillance footage), so learned normal/abnormal priors may not transfer to deployment contexts with substantially different cultural norms or visual environments without careful bias auditing.

\end{document}